# Communication-efficient Distributed Sparse Linear Discriminant Analysis

Lu Tian* and Quanquan Gu†

## Abstract

We propose a communication-efficient distributed estimation method for sparse linear discriminant analysis (LDA) in the high dimensional regime. Our method distributes the data of size $N$ into $m$ machines, and estimates a local sparse LDA estimator on each machine using the data subset of size $N/m$. After the distributed estimation, our method aggregates the debiased local estimators from $m$ machines, and sparsifies the aggregated estimator. We show that the aggregated estimator attains the same statistical rate as the centralized estimation method, as long as the number of machines $m$ is chosen appropriately. Moreover, we prove that our method can attain the model selection consistency under a milder condition than the centralized method. Experiments on both synthetic and real datasets corroborate our theory.

## 1 Introduction

High dimensionality is a frequently confronted problem in many applications of machine learning. It increases time and space requirements for processing the data. Moreover, many machine learning methods tend to over-fit and become less interpretable in the presence of many irrelevant or redundant features. A common way to address this problem is the dimensionality reduction. Principal Component Analysis (PCA) (Jolliffe, 2002) is probably the most widely used dimensionality reduction method. However, it is an unsupervised dimensionality reduction method and does not consider the labels of the data. In order to take the label information into account, supervised dimensionality reduction methods are favored. Linear Discriminant Analysis (LDA) (Anderson, 1968), which is initially proposed as a classification method, is an important supervised dimensionality reduction method. Let $\boldsymbol{X}$ and $\boldsymbol{Y}$ be two $d$-dimensional random vectors following two normal distributions, $\boldsymbol{X} \sim N(\boldsymbol{\mu}_1, \boldsymbol{\Sigma}^*)$ and $\boldsymbol{Y} \sim N(\boldsymbol{\mu}_2, \boldsymbol{\Sigma}^*)$, which share the same covariance matrix $\boldsymbol{\Sigma}^*$ but with different mean vectors $\boldsymbol{\mu}_1$ and $\boldsymbol{\mu}_2$. For a new observation $\boldsymbol{Z}$ that is drawn with equal prior probability from the two normal distributions, the Fisher's linear discriminant rule takes the form

$$\psi(\boldsymbol{Z}) = \mathbb{1}((\boldsymbol{Z} - \boldsymbol{\mu})^\top \boldsymbol{\Theta}^* \boldsymbol{\mu}_d > 0), \tag{1.1}$$

---

*Department of Systems and Information Engineering, University of Virginia, Charlottesville, VA 22904, USA; e-mail: `lt2eu@virginia.edu`

†Department of Systems and Information Engineering, Department of Computer Science, University of Virginia, Charlottesville, VA 22904, USA; e-mail: `qg5w@virginia.edu`

where $\boldsymbol{\mu} = (\boldsymbol{\mu}_1 + \boldsymbol{\mu}_2)/2$, $\boldsymbol{\mu}_d = \boldsymbol{\mu}_1 - \boldsymbol{\mu}_2$, $\boldsymbol{\Theta}^* = \boldsymbol{\Sigma}^{*-1}$ is the precision matrix (a.k.a., the inverse covariance matrix), and $\mathbb{1}(\cdot)$ is the indicator function. It is well known that the Fisher's linear discriminant rule minimizes the misclassification rate and it is Bayesian optimal. In practice, $\boldsymbol{\mu}_1, \boldsymbol{\mu}_2$ and $\boldsymbol{\Sigma}^*$ are unknown, and we need to estimate $\boldsymbol{\mu}_1$, $\boldsymbol{\mu}_2$ and $\boldsymbol{\Sigma}^*$ from observations. More specifically, let $\{\boldsymbol{X}_i : 1 \leq i \leq n_1\}$ and $\{\boldsymbol{Y}_i : 1 \leq i \leq n_2\}$ be independently and identically distributed random samples from $N(\boldsymbol{\mu}_1, \boldsymbol{\Sigma}^*)$ and $N(\boldsymbol{\mu}_2, \boldsymbol{\Sigma}^*)$ respectively. The classical estimations of $\boldsymbol{\mu}_1, \boldsymbol{\mu}_2$ and $\boldsymbol{\Theta}^*$ in the classical regime are the sample means $\widehat{\boldsymbol{\mu}}_1 = n_1^{-1} \sum_{i=1}^{n_1} \boldsymbol{X}_i$ and $\widehat{\boldsymbol{\mu}}_2 = n_2^{-1} \sum_{i=1}^{n_2} \boldsymbol{Y}_i$, and $\widehat{\boldsymbol{\Theta}} = \widehat{\boldsymbol{\Sigma}}^{-1}$, where $\widehat{\boldsymbol{\Sigma}} = n^{-1} \big[ \sum_{i=1}^{n_1} (\boldsymbol{X}_i - \widehat{\boldsymbol{\mu}}_1)(\boldsymbol{X}_i - \widehat{\boldsymbol{\mu}}_1)^\top + \sum_{i=1}^{n_2} (\boldsymbol{Y}_i - \widehat{\boldsymbol{\mu}}_2)(\boldsymbol{Y}_i - \widehat{\boldsymbol{\mu}}_2)^\top \big]$ is the pooled sample covariance matrix with $n = n_1 + n_2$. Plugging these estimators into (1.1) gives rise to the empirical version of $\psi(\boldsymbol{Z})$, i.e., $\widehat{\psi}(\boldsymbol{Z})$. Theoretical properties of $\widehat{\psi}(\boldsymbol{Z})$ have been well studied when $d$ is fixed, e.g., see Anderson (1968). However, in the high-dimensional regime where $d$ increases as $n$, the pooled sample covariance matrix procedure is not well-conditioned and the plug-in estimator is not reliable. For example, Bickel and Levina (2004) showed that it is asymptotically equivalent to random guess when the dimensionality increases at some rate comparable to the number of samples. To overcome this curse of dimensionality, it is natural to impose some structural assumptions on the parameters of the discriminant rule in (1.1). For example, Cai and Liu (2011) made the assumption that $\boldsymbol{\beta}^* = \boldsymbol{\Theta}^* \boldsymbol{\mu}_d$ is a sparse vector and proposed the following estimator:

$$\widehat{\boldsymbol{\beta}} = \operatorname*{argmin}_{\boldsymbol{\beta}} \|\boldsymbol{\beta}\|_1 \quad \text{subject to} \quad \|\widehat{\boldsymbol{\Sigma}} \boldsymbol{\beta} - (\widehat{\boldsymbol{\mu}}_1 - \widehat{\boldsymbol{\mu}}_2)\|_\infty \leq \lambda, \tag{1.2}$$

where $\|\boldsymbol{\beta}\|_1 = \sum_{j=1}^d |\beta_j|$ is the $\ell_1$ norm, and $\|\cdot\|_\infty$ is the element-wise max norm, $\widehat{\boldsymbol{\Sigma}}$, $\widehat{\boldsymbol{\mu}}_1$ and $\widehat{\boldsymbol{\mu}}_2$ are defined as above and $\lambda > 0$ is a tuning parameter. In our study, we will focus on the above sparse LDA estimator, because it is comparable to or even better than many other sparse LDA estimators (Shao et al., 2011; Mai et al., 2012; Fan et al., 2012).

On the other hand, with the increase in the volume of data used for machine learning, and the availability of distributed computing resources, distributed statistical estimation (Mcdonald et al., 2009; Balcan et al., 2012; Zhang et al., 2012, 2013; Rosenblatt and Nadler, 2014; Lee et al., 2015; Battey et al., 2015) and distributed optimization (Zinkevich et al., 2010; Boyd et al., 2011; Dekel et al., 2012) have received increasing attention. The main bottleneck in distributed computing is usually the communication between machines, so the overarching goal of the algorithm design in distributed setting is to reduce the communication costs, while trying to achieve comparable performance as centralized algorithms. The problem becomes even more challenging when high dimensionality meets huge data size.

To address the challenge of both high dimensionality and huge data size, in this paper, we propose a distributed sparse linear discriminant analysis method. In the proposed algorithm, each "worker" machine generates a local estimator for the sparse LDA and sends it to the "master" machine, where all local estimators are averaged to form an aggregated estimator. At the core of our algorithm is an unbiased estimator for the sparse linear discriminant analysis. It is worth noting that our proposed algorithm requires only one round of communication between the worker nodes and the master node. That is, each worker machine only needs to send a vector to the master node. Thus, our algorithm is very communication-efficient. We prove the estimation error bounds for the proposed algorithm in terms of different norms. More specifically, we show that the proposed distributed algorithm attains $O(\sqrt{s \log d / N} + \max(s, s') m \sqrt{s} \log d / N)$ estimation error bound in terms of $\ell_2$ norm, where $N$ is the total sample size, $m$ is the number of machines, $d$ is the dimensionality, $s = \|\boldsymbol{\beta}^*\|_0$ and $s' = \max_{1 \leq j \leq d} \|\boldsymbol{\theta}_j^*\|_0$ are the number of nonzero elements in $\boldsymbol{\beta}^*$ and $\boldsymbol{\theta}_j^*$ respectively, with $\boldsymbol{\theta}_j^*$ being the $j$-th column of $\boldsymbol{\Theta}^*$. From the estimation error bound, we address an important question that how to choose $m$ such that the information loss due to the data parallelism is negligible. In particular, if the machine number $m$ satisfies $m \lesssim \sqrt{N/\log d}/\max(s, s')$, our distributed algorithm attains the same statistical rate as the centralized estimator (Cai and Liu, 2011), which is $O(\sqrt{s \log d/N})$ in terms of $\ell_2$-norm. Furthermore, we show that given $\min_j |\beta_j^*| \gtrsim \sqrt{\log d/N}$, our estimator achieves the model selection consistency, which matches the minimax lower bound for support recovery in sparse LDA (Fan et al., 2012; Kolar and Liu, 2015). However, the model selection consistency established in Kolar and Liu (2015) relies on the irrepresentable condition, which is very stringent. In sharp contrast, the model selection consistency of our algorithm does not need this condition.

**Notation** We summarize here the notations to be used throughout the paper. We use lowercase letters $x, y, \ldots$ to denote scalars, bold lowercase letters $\mathbf{x}, \mathbf{y}, \ldots$ for vectors, and bold uppercase letters $\mathbf{X}, \mathbf{Y}, \ldots$ for matrices. We denote random vectors by $\boldsymbol{X}, \boldsymbol{Y}$. We denote $\mathbf{e}_j$ as the column vector whose $j$-th entry is one and others are zeros. Let $\mathbf{A} = [A_{ij}] \in \mathbb{R}^{d \times d}$ be a $d \times d$ matrix and $\mathbf{x} = [x_1, \ldots, x_d]^\top \in \mathbb{R}^d$ be a $d$-dimensional vector. For $0 < q < \infty$, we define the $\ell_0$, $\ell_q$ and $\ell_\infty$ vector norms as $\|\mathbf{x}\|_0 = \sum_{i=1}^d \mathbb{1}(x_i \neq 0), \|\mathbf{x}\|_q = (\sum_{i=1}^d |x_i|^q)^{1/q}, \|\mathbf{x}\|_\infty = \max_{1 \leq i \leq d} |x_i|$, where $\mathbb{1}(\cdot)$ represents the indicator function. For any real number $C$ and symmetric matrix $\mathbf{A}$, $\mathbf{A} \succ C$ means that the minimum eigenvalue of $\mathbf{A}$ is larger than $C$. Specifically, $\mathbf{A} \succ 0$ means that $\mathbf{A}$ is a positive definite matrix. We use the following notation for the matrix $\ell_\infty$, $\ell_1$, $\ell_{\infty,\infty}$ and $\ell_{1,1}$ norms:

$$\|\mathbf{A}\|_\infty = \max_{1 \leq j \leq d} \sum_{k=1}^d |A_{jk}|, \quad \|\mathbf{A}\|_1 = \|\mathbf{A}^\top\|_\infty, \quad \|\mathbf{A}\|_{\infty,\infty} = \max_{1 \leq i,j \leq d} |A_{ij}|, \quad \|\mathbf{A}\|_{1,1} = \sum_{1 \leq i,j \leq d} |A_{ij}|.$$

For a vector $\mathbf{x}$ and an index set $S$, $\mathbf{x}_S$ denotes the vector such that $[\mathbf{x}_S]_j = x_j$ if $j \in S$, and $[\mathbf{x}_S]_j = 0$ otherwise. For sequences $f_n, g_n$, we write $f_n = O(g_n)$ if $|f_n| \leq C|g_n|$ for some $C > 0$ independent of $n$ and all $n > D$, where $D$ is a positive integer. We also make use of the notation $f_n \lesssim g_n$ ($f_n \gtrsim g_n$) if $f_n$ is less than (greater than) $g_n$ up to a constant. In this paper, $C, c, C', C_1$ etc. denote various absolute constants, not necessarily the same at each occurrence.

## 2 Related Work

In this section, we briefly review the related work on sparse linear discriminant analysis (LDA) and distributed estimation.

LDA has been widely studied in the high dimensional regime where the number of features $d$ can increase as the sample size $n$ (Shao et al., 2011; Cai and Liu, 2011; Mai et al., 2012; Fan et al., 2012). One important problem in the high dimensional regime is that the estimation of $\boldsymbol{\Theta}^*$ will be unstable because the sample covariance matrix $\widehat{\boldsymbol{\Sigma}}$ is often singular. To address this problem, a common assumption is that both $\boldsymbol{\mu}_d$ and $\boldsymbol{\Sigma}^*$ are sparse. Under this assumption, Shao et al. (2011) proposed to use a thresholding procedure to estimate $\boldsymbol{\mu}_d$ and $\boldsymbol{\Sigma}^*$ respectively, followed by the standard procedure to estimate $\psi(\boldsymbol{Z})$. Cai and Liu (2011) assumed that $\boldsymbol{\beta}^* = \boldsymbol{\Theta}^* \boldsymbol{\mu}_d$ is sparse and estimated it directly. While sparse LDA has been investigated extensively, it is not clear how to extend it to the distributed setting, where the data are distributed on multiple machines.

With the growth of the size of available datasets, distributed algorithms become more and more attractive in the machine learning and optimization communities. In general, distributed algorithm can be categorized into two families: (1) data parallelism, which distributes the data across different parallel computing nodes/machines; and (2) task parallelism, which distributes tasks performed by threads across different parallel computing nodes. In this paper, we focus on data parallelism. The most important problem in data parallelism is how to minimize the communication cost among different machines. A commonly used approach in distributed statistical estimation is averaging: each "worker" machine generates a local estimator and sends it to the "master" machine where all local estimators are averaged to form an aggregated estimator. This type of approach has been first studied by Mcdonald et al. (2009); Zinkevich et al. (2010); Zhang et al. (2012, 2013); Balcan et al. (2012). Nevertheless, the above distributed statistical estimation methods are in the classical regime. In the high dimensional regime, averaging is not an effective way for aggregation (Rosenblatt and Nadler, 2014). Moreover, many estimators in the high dimensional regime are based on the penalized estimation, which introduces some bias to the estimator. For example, the Lasso estimator (Tibshirani, 1996) is biased due to the $\ell_1$-norm penalty. Since averaging only reduces variances, not the bias, the performance of averaged estimator is no better than the local estimator due to the aggregation of bias when averaging. To address this problem, Lee et al. (2015) proposed distributed sparse regression methods, which exploits the debiased estimators proposed in Javanmard and Montanari (2014); Van de Geer et al. (2014) for distributed sparse regression. Similar distributed regression methods are proposed by Battey et al. (2015) for both distributed statistical estimation and hypothesis testing. However, all the above studies on distributed statistical estimation are focused on regression. It is not easy to extend them to distributed dimensionality reduction.

In fact, the problem of distributed dimensionality reduction is still relatively under-studied. Liang et al. (2014) proposed a distributed approximate PCA algorithm, which speeds up the computation and needs low communication cost but with a low accuracy loss. Balcan et al. (2015) extended the kernel PCA to the distributed setting and proposed a communication-efficient distributed kernel PCA algorithm. Valcarcel Macua et al. (2011) developed a distributed algorithm for linear discriminant analysis on a single-hop network. Nevertheless, all these algorithms are in the classical regime, and cannot be applied to sparse LDA in the high dimensional regime.

## 3 Distributed Sparse Linear Discriminant Analysis

In this section, we present a distributed linear discriminant analysis algorithm.

The problem setup of distributed sparse linear discriminant analysis is as follows: Let $\mathbf{X}^{(l)} \in \mathbb{R}^{n_{1l} \times d}, l \in \{1, 2, \ldots, m\}$ be the data matrix of the first class stored on the $l$-th machine, each row of which is sampled i.i.d. from the multivariate normal distribution $N(\boldsymbol{\mu}_1, \boldsymbol{\Sigma}^*)$. Similarly, let $\mathbf{Y}^{(l)} \in \mathbb{R}^{n_{2l} \times d}, l \in \{1, 2, \ldots, m\}$ be the data matrix of the second class stored on the $l$-th machine, where each row is sampled i.i.d. from the multivariate normal distribution $N(\boldsymbol{\mu}_2, \boldsymbol{\Sigma}^*)$. Without loss of generality, we assume $n_{11} = n_{12} = \ldots = n_{1m} = n_1$ and $n_{21} = n_{22} = \ldots = n_{2m} = n_2$. Let $n = n_1 + n_2$, which is the total number of data stored in a single machine. We also assume $n_1 \leq n_2$ and $n_1 = rn$, where $r \leq 1/2$ is a constant. We propose a distributed sparse LDA algorithm based on Cai and Liu (2011) to directly estimate $\boldsymbol{\beta}^*$ in Algorithm 1.

In detail, for the $l$-th machine, we denote by $\boldsymbol{X}_i^{(l)}$ and $\boldsymbol{Y}_i^{(l)}$ the $i$-th row of $\mathbf{X}^{(l)}$ and $\mathbf{Y}^{(l)}$ respectively. On each machine, we can use the sparse LDA estimator in (1.2) to obtain a local

**Algorithm 1** Distributed Sparse Linear Discriminant Analysis
___
**Require:** $\mathbf{X}^{(1)}, \ldots, \mathbf{X}^{(m)}, \mathbf{Y}^{(1)}, \ldots, \mathbf{Y}^{(m)}$
**Ensure:** $\bar{\boldsymbol{\beta}}$, the aggregated sparse discriminant vector
  **Workers:**
  Each worker computes $\widehat{\boldsymbol{\Sigma}}^{(l)}$ and $\widehat{\boldsymbol{\mu}}_1^{(l)}, \widehat{\boldsymbol{\mu}}_2^{(l)}$
  Each worker computes a local sparse LDA estimator $\widehat{\boldsymbol{\beta}}^{(l)}$ by (3.1)
  Each worker computes a debiased estimator $\widetilde{\boldsymbol{\beta}}^{(l)}$ by (3.4)
  Each worker sends $\widetilde{\boldsymbol{\beta}}^{(l)}$ to the master machine
  **Master:**
  **while** waiting for $\widetilde{\boldsymbol{\beta}}^{(l)}$ sent from all workers **do**
    **if** received $\widetilde{\boldsymbol{\beta}}^{(l)}$ from all workers **then**
      Compute the aggregated sparse estimator $\bar{\boldsymbol{\beta}}$ by (3.5)
    **end if**
  **end while**
___

estimator as the following:

$$\widehat{\boldsymbol{\beta}}^{(l)} = \underset{\boldsymbol{\beta} \in \mathbb{R}^d}{\operatorname{argmin}} \|\boldsymbol{\beta}\|_1 \quad \text{subject to} \quad \left\|\widehat{\boldsymbol{\Sigma}}^{(l)}\boldsymbol{\beta} - \widehat{\boldsymbol{\mu}}_d^{(l)}\right\|_\infty \leq \lambda, \tag{3.1}$$

where $\lambda > 0$ is a tuning parameter, $\widehat{\boldsymbol{\mu}}_d^{(l)} = \widehat{\boldsymbol{\mu}}_1^{(l)} - \widehat{\boldsymbol{\mu}}_2^{(l)}$ with sample means $\widehat{\boldsymbol{\mu}}_1^{(l)} = (\sum_{i=1}^{n_1} \boldsymbol{X}_i^{(l)})/n_1$ and $\widehat{\boldsymbol{\mu}}_2^{(l)} = (\sum_{i=1}^{n_2} \boldsymbol{Y}_i^{(l)})/n_2$ and

$$\widehat{\boldsymbol{\Sigma}}^{(l)} = \frac{1}{n}\bigg[\sum_{i=1}^{n_1}(\boldsymbol{X}_i^{(l)} - \widehat{\boldsymbol{\mu}}_1^{(l)})(\boldsymbol{X}_i^{(l)} - \widehat{\boldsymbol{\mu}}_1^{(l)})^\top + \sum_{i=1}^{n_2}(\boldsymbol{Y}_i^{(l)} - \widehat{\boldsymbol{\mu}}_2^{(l)})(\boldsymbol{Y}_i^{(l)} - \widehat{\boldsymbol{\mu}}_2^{(l)})^\top\bigg],$$

which is the total intra-class sample covariance matrix of the $l$-th machine. The choice of $\lambda$ will be discussed in Section 4.

The estimator in (3.1) is biased due to the shrinkage property of the estimator. Since averaging only reduce the variance, rather than the bias, if we naively average $\widehat{\boldsymbol{\beta}}^{(l)}$'s, the error bound of the averaged estimator will remain in the same order as that of the local estimators. To address the bias, several debiasing techniques have been proposed, such as Lee et al. (2015) and Battey et al. (2015). However, Lee et al. (2015) focused on the Lasso estimator, and the debiasing method proposed in Battey et al. (2015) is only suitable for regularized estimators. In order to construct an unbiased estimator for the Dantzig-type estimator, we propose a new debiasing procedure as follows: First, the CLIME estimator (Cai et al., 2011) is used to estimate the precision matrix:

$$\widehat{\boldsymbol{\Theta}}^{(l)} = \operatorname{argmin} \|\boldsymbol{\Theta}\|_{1,1} \quad \text{subject to} \quad \|\boldsymbol{\Theta}^\top \widehat{\boldsymbol{\Sigma}}^{(l)} - \mathbf{I}\|_{\infty,\infty} \leq \lambda', \tag{3.2}$$

where $\lambda'$ is a tuning parameter, and its choice will be clear from Section 4. It is worth noting that (3.2) can be decomposed into $d$ independent optimization problems, where each one takes the form

$$\widehat{\boldsymbol{\theta}}_j^{(l)} = \operatorname{argmin} \|\boldsymbol{\theta}\|_1 \quad \text{subject to} \quad \|\widehat{\boldsymbol{\Sigma}}^{(l)}\boldsymbol{\theta} - \mathbf{e}_j\|_\infty \leq \lambda', \tag{3.3}$$

for $j \in \{1, 2, \ldots, d\}$ and $\widehat{\boldsymbol{\theta}}_j^{(l)}$ corresponds to the $j$-th column of $\widehat{\boldsymbol{\Theta}}^{(l)}$. Therefore, they can be solved in parallel.

Second, based on $\widehat{\boldsymbol{\Theta}}^{(l)}$, we construct a debiased estimator $\widetilde{\boldsymbol{\beta}}^{(l)}$ in the following way:

$$\widetilde{\boldsymbol{\beta}}^{(l)} = \widehat{\boldsymbol{\beta}}^{(l)} - \widehat{\boldsymbol{\Theta}}^{(l)\top}\left(\widehat{\boldsymbol{\Sigma}}^{(l)}\widehat{\boldsymbol{\beta}}^{(l)} - \widehat{\boldsymbol{\mu}}_d^{(l)}\right). \tag{3.4}$$

Note that the second term in the right hand side of (3.4) can be seen as the estimation of the bias introduced by the penalized estimator in (3.2). We subtract the estimation of the bias from $\widehat{\boldsymbol{\beta}}^{(l)}$ and obtain an unbiased estimator $\widetilde{\boldsymbol{\beta}}^{(l)}$.

Finally, the workers send back the unbiased local estimators in (3.4) generated by different machines to the master node, and the master node averages all the debiased local estimators followed by hard thresholding in order to get a sparse estimator. More specifically, the sparse aggregated estimator is as follows

$$\bar{\boldsymbol{\beta}} = \text{HT}\left(\frac{1}{m}\sum_{l=1}^m \widetilde{\boldsymbol{\beta}}^{(l)}, t\right), \tag{3.5}$$

where $\text{HT}(\cdot)$ is the hard thresholding operator, which is defined as

$$[\text{HT}(\boldsymbol{\beta}, t)]_j = \begin{cases} \beta_j, & \text{if } |\beta_j| > t, \\ 0, & \text{if } |\beta_j| \leq t. \end{cases}$$

Here $t > 0$ is a pre-specified threshold. The setting of $t$ will be discussed in Section 4.

The proposed distributed algorithm has a low communication cost. In detail, compared with the naive distributed algorithm in which $\widehat{\boldsymbol{\Sigma}}^{(l)}$'s and $\widehat{\boldsymbol{\mu}}_d^{(l)}$'s are computed separately on each machine and then sent back to the master node, our algorithm only needs to send $d$-dimensional vectors rather than $d \times d$ matrices to the master node, which significantly reduces the communication cost. Moreover, we will prove later that, while keeping low communication cost, our algorithm can attain the same convergence rate as the centralized method if we choose the number of machines appropriately.

The time complexity of our algorithm can be illustrated as follows: in order to obtain $\widehat{\boldsymbol{\beta}}^{(l)}$, the main computation overhead lies on computing $\widehat{\boldsymbol{\Sigma}}^{(l)}$, whose time complexity is $O(Nd^2/m)$. For the CLIME estimator, using the FastCLIME method (Pang et al., 2014), the time complexity is $O(d^2)$. Thus the total time complexity of the proposed algorithm per machine is $O(Nd^2/m)$. In contrast, for centralized estimator which collects the data from all local machines and performs the estimation, the time complexity is $O(Nd^2)$. Therefore, as the number of machine grows, a near linear speedup in the number of machines can be achieved for our distributed algorithm. Furthermore, as will be demonstrated in the main theory, in order to make the information loss caused by the data parallelism negligible, the appropriate choice of $m$ can be as large as $O(\sqrt{N})$, which implies a time complexity of $O(d^2\sqrt{N})$ on each machine. This suggests that the proposed algorithm has a lower time complexity while attaining the same statistical rate as the centralized method.

## 4 Main Theory

In this section, we establish the main theory for our distributed LDA algorithm. Before we present the main result of this paper, we first lay out some necessary assumptions.

We make the following assumptions on the covariance matrix and the precision matrix of the two normal distributions.

**Assumption 4.1.** *There exists a constant $K \geq 1$, such that the maximum and minimal eigenvalues of $\boldsymbol{\Sigma}^*$ can be bounded as follows:*

$$1/K \leq \lambda_{\min}(\boldsymbol{\Sigma}^*) \leq \lambda_{\max}(\boldsymbol{\Sigma}^*) \leq K.$$

*Furthermore we assume that $K$ does not increase as $d$ goes to infinity.*

**Assumption 4.2.** *$\boldsymbol{\Theta}^*$ belongs to the following set:*

$$\mathcal{U}(s', M) = \Big\{\boldsymbol{\Theta} : \boldsymbol{\Theta} \succ 0, \|\boldsymbol{\Theta}\|_1 \leq M, \max_{1 \leq j \leq d} \sum_{k=1}^{d} \mathbb{1}(\Theta_{jk} \neq 0) \leq s'\Big\}.$$

Assumption 4.2 is a common assumption made in the literature of sparse precision matrix estimation (Cai et al., 2011). It implies that the data can be viewed as generated from a sparse Gaussian graphical model, where the maximum degree of the graph is no larger than $s'$. Note that Assumption 4.2 immediately implies that $\|\boldsymbol{\theta}_j^*\|_1 \leq \|\boldsymbol{\Theta}^*\|_1 \leq M$ for all $j \in \{1, 2, \ldots, d\}$.

In most literatures on high dimensional sparse estimation (Bickel et al., 2009; Negahban et al., 2009), it is assumed that the sample covariance matrix satisfies the restricted eigenvalue condition. Following is the definition of the restricted eigenvalue condition that we use in this paper.

**Definition 4.3.** *A matrix $\mathbf{A} \in \mathbb{R}^{d \times d}$ satisfies the restricted eigenvalue (RE) condition with parameters $(s, \alpha, \gamma)$ if and only if for any index set $S$ with $|S| \leq s$, for any vector $\mathbf{v}$ in the cone*

$$\mathbb{C}(S, \alpha) = \{\mathbf{v} \in \mathbb{R}^d : \|\mathbf{v}_{S^c}\|_1 \leq \alpha \|\mathbf{v}_S\|_1\},$$

*we have $\mathbf{v}^\top \mathbf{A} \mathbf{v} \geq \gamma \|\mathbf{v}\|_2^2$.*

With this definition, the assumption made on the sample covariance matrices can be presented as follows.

**Condition 4.4.** *For each $l \in \{1, 2, \ldots, m\}$, $\widehat{\boldsymbol{\Sigma}}^{(l)}$ satisfies the restricted eigenvalue condition with respect to the parameters $(\max\{s, s'\}, 1, \lambda_{\min}(\boldsymbol{\Sigma}^*)/16)$.*

The following proposition shows that Condition 4.4 is satisfied with high probability when the sample size $n$ is sufficiently large.

**Proposition 4.5.** *If $n > \max\{s, s'\} r^{-1} C_1 K^3 \log d$, Condition 4.4 is satisfied with probability at least $1 - mC_2 \exp(-C_3 n) - 2m/d$, where $C_1$, $C_2$ and $C_3$ are absolute constants.*

Now we are ready to present the main theorem bounding the estimation error of $\bar{\boldsymbol{\beta}}$.

**Theorem 4.6.** *Under Assumptions 4.1, 4.2 and Condition 4.4, if $\lambda = C_1 K^2 \sqrt{\log d/(rn)} \|\boldsymbol{\beta}^*\|_1$, $\lambda' = C_2 K^2 M \sqrt{\log d/n}$ for some $C_1$ and $C_2$, and $t$ is chosen as*

$$t = C' M \sqrt{\frac{\log d}{N}} \|\boldsymbol{\beta}^*\|_1 + C'' \max(s, s') M \frac{m \log d}{N} \|\boldsymbol{\beta}^*\|_1, \tag{4.1}$$

where $C'$ and $C''$ are absolute constants, then the following inequality holds with probability at least $1 - 18m/d - 4/d$:

$$\|\bar{\boldsymbol{\beta}} - \boldsymbol{\beta}^*\|_\infty \leq C'M\sqrt{\frac{\log d}{N}}\|\boldsymbol{\beta}^*\|_1 + C'' \max(s, s')M\frac{m \log d}{N}\|\boldsymbol{\beta}^*\|_1. \tag{4.2}$$

Moreover, with probability at least $1 - 18m/d - 4/d$ we have

$$\|\bar{\boldsymbol{\beta}} - \boldsymbol{\beta}^*\|_2 \leq \sqrt{s}C'M\sqrt{\frac{\log d}{N}}\|\boldsymbol{\beta}^*\|_1 + \sqrt{s}C'' \max(s, s')M\frac{m \log d}{N}\|\boldsymbol{\beta}^*\|_1, \tag{4.3}$$

and with probability at least $1 - 18m/d - 4/d$ we have

$$\|\bar{\boldsymbol{\beta}} - \boldsymbol{\beta}^*\|_1 \leq sC'M\sqrt{\frac{\log d}{N}}\|\boldsymbol{\beta}^*\|_1 + sC'' \max(s, s')M\frac{m \log d}{N}\|\boldsymbol{\beta}^*\|_1. \tag{4.4}$$

The proof of Theorem 4.6 is in Appendix A. It is worth noting that in the linear discriminant analysis, only the direction of $\bar{\boldsymbol{\beta}}$ affects the discrimination, while the norm of $\bar{\boldsymbol{\beta}}$ does not matter. Therefore, the relative error, i.e., the ratio of the norm of $\bar{\boldsymbol{\beta}} - \boldsymbol{\beta}^*$ to the norm of $\boldsymbol{\beta}^*$, can better characterize the accuracy of the estimator.

**Remark 4.7.** *The centralized estimator can be regarded as a special case of the biased estimator (3.1) where $m = 1$ and $n = N$. Hence by Lemma B.4 the error bound of the centralized estimator can be obtained: with probability at least $1 - 6/d$ we have*

$$\|\widehat{\boldsymbol{\beta}}^{\text{centralized}} - \boldsymbol{\beta}^*\|_1 \leq sCK^2\sqrt{\frac{\log d}{N}}\|\boldsymbol{\beta}^*\|_1,$$

*where $C$ is a constant. Compared with our distributed estimator, it can be seen that the error bound of the centralized estimator is of the same order with the first term of our proposed estimator, which is in the order of $O(\sqrt{\log d/N})$. And the second term of the error bound of our estimator is in the order of $O(m \log d/N)$, reflecting the loss caused by the data distribution and one round of communication.*

**Corollary 4.8.** *Under the same assumptions with Theorem 4.6, if the number of machines $m$ is chosen to be*

$$m \lesssim \frac{1}{\max(s, s')}\sqrt{\frac{N}{\log d}}, \tag{4.5}$$

*then with probability at least $1 - 18m/d - 4/d$ the following inequalities holds:*

$$\|\bar{\boldsymbol{\beta}} - \boldsymbol{\beta}^*\|_\infty \leq CM\sqrt{\frac{\log d}{N}}\|\boldsymbol{\beta}^*\|_1, \qquad \|\bar{\boldsymbol{\beta}} - \boldsymbol{\beta}^*\|_2 \leq \sqrt{s}CM\sqrt{\frac{\log d}{N}}\|\boldsymbol{\beta}^*\|_1,$$

$$\|\bar{\boldsymbol{\beta}} - \boldsymbol{\beta}^*\|_1 \leq sCM\sqrt{\frac{\log d}{N}}\|\boldsymbol{\beta}^*\|_1,$$

*where $C$ is a constant.*

**Remark 4.9.** *Generally speaking, the distributed estimation may cause information loss and lead to a worse estimation error bound. However, Corollary 4.8 suggests that if the number of machines m satisfies $m \lesssim \sqrt{N/\log d}/\max(s,s')$ when $N, d, s$ and $s'$ grow, the information loss is negligible and the distributed algorithm can attain the same rate of convergence as the centralized algorithm.*

In fact, the $\ell_\infty$ estimation error bound in Theorem 4.6 ensures that the estimated parameter vector correctly excludes all non-informative variables and includes all useful variables provided that

$$|\beta_j^*| > C'M\sqrt{\frac{\log d}{N}}\|\boldsymbol{\beta}^*\|_1 + C''\max(s,s')M\frac{m\log d}{N}\|\boldsymbol{\beta}^*\|_1,$$

where $C'$ and $C''$ are the same as in Theorem 4.6. Therefore, in order to achieve the model selection consistency, it is sufficient to assume that the minimum signal strength $\beta_{\min} := \min_{j \in S}|\beta_j^*|$ is not too small. More specifically, we have the following theorem:

**Theorem 4.10.** *Under the same assumptions with Theorem 4.6, if*

$$\beta_{\min} > C'M\sqrt{\frac{\log d}{N}}\|\boldsymbol{\beta}^*\|_1 + C''\max(s,s')M\frac{m\log d}{N}\|\boldsymbol{\beta}^*\|_1, \tag{4.6}$$

*where $C'$ and $C''$ are those appeared in Theorem 4.6, we have with probability higher than $1 - 18m/d - 4/d$ that $\operatorname{sign}(\bar\beta_j) = \operatorname{sign}(\beta_j^*)$ for any $j \in \{1, 2, \ldots, d\}$.*

Similar to Corollary 4.8, we have the following conclusion:

**Corollary 4.11.** *Under the same assumptions with Theorem 4.10, if the following two condition holds:*

$$m \lesssim \frac{1}{\max(s,s')}\sqrt{\frac{N}{\log d}}, \quad \beta_{\min} > CM\sqrt{\frac{\log d}{N}}\|\boldsymbol{\beta}^*\|_1 \tag{4.7}$$

*for some $C$, then we have with probability at least $1 - 18m/d - 4/d$ that $\operatorname{sign}(\bar\beta_j) = \operatorname{sign}(\beta_j^*)$ for any $j \in \{1, 2, \ldots, d\}$.*

**Remark 4.12.** *In Cai and Liu (2011), the authors did not provide theoretical guarantee on the support recovery. Mai et al. (2012) showed that the condition on $\beta_{\min}$ needed for model selection consistency is $\beta_{\min} \gtrsim s\sqrt{\log(sd)/N}$. The condition for the ROAD estimator proposed in Fan et al. (2012) to satisfy the model selection consistency is $\beta_{\min} \gtrsim \sqrt{\log d/N}$ (Kolar and Liu, 2015), which is proved to be minimax optimal. It is obvious that our condition implied by Corollary 4.11 matches the minimax lower bound in Kolar and Liu (2015) and is better than Mai et al. (2012). However, for the ROAD estimator, a very stringent irrepresentable condition is required for the model selection consistency to hold. For our algorithm, the irrepresentable condition is not required.*

## 5 Experiments

In this section, we verify the performance of the distributed LDA algorithm using both synthetic data and real data. We compared it with the centralized sparse LDA estimator, and naively averaged sparse LDA estimator. In the centralized SLDA, all samples are collected in one machine and the model is estimated by Cai and Liu (2011). In the naively averaged SLDA estimator, we apply Cai and Liu (2011) to the data on each machine to obtain local estimators. The local estimators are directly averaged without debiasing. In other words, the naively averaged SLDA estimator can be written as $\widehat{\boldsymbol{\beta}}_{\mathrm{n}} = (\sum_{l=1}^{m} \widehat{\boldsymbol{\beta}}^{(l)})/m$.

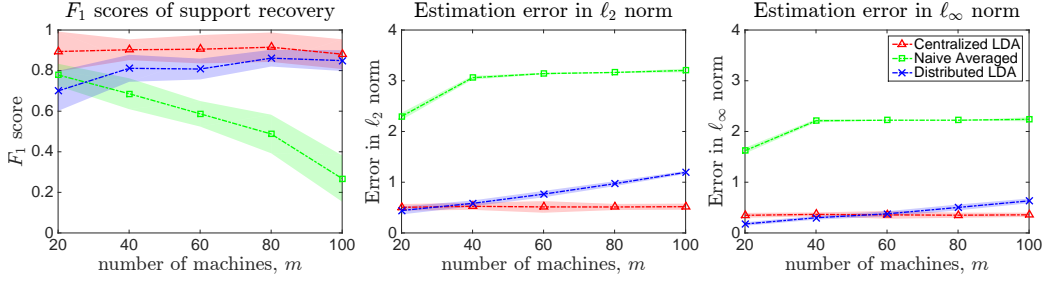

**Figure 1:** The $F_1$ score and estimation error (in $\ell_2$ and $\ell_\infty$ norms) of the proposed estimator versus the centralized estimator and the naive averaged estimator when the total sample size $N$ is fixed as 10000.

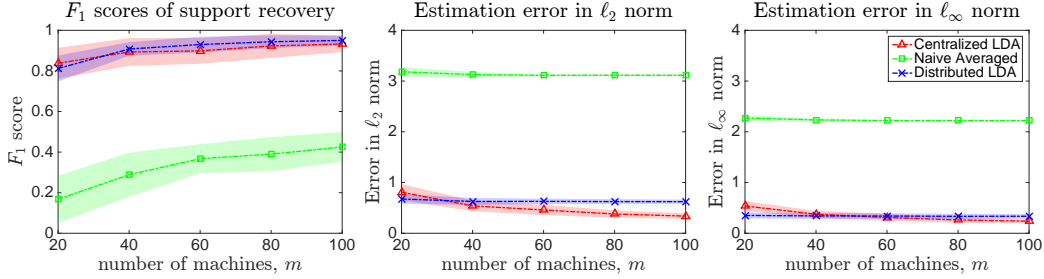

**Figure 2:** The $F_1$ score and estimation error (in $\ell_2$ and $\ell_\infty$ norms) of the proposed estimator versus the centralized estimator and the naive averaged estimator when the sample size on each machine $n$ is set to 200.

### 5.1 Synthetic Data Experiments

The synthetic data are generated by setting $\boldsymbol{\Sigma}^*$ and $\boldsymbol{\mu}_1, \boldsymbol{\mu}_2$ as follows: $\boldsymbol{\Sigma}^* \in \mathbb{R}^{d \times d}$ with $d = 200$, and $\Sigma_{jk}^* = 0.8^{|j-k|}$ for all $j, k \in \{1, \ldots, d\}$. Additionally, we choose $\boldsymbol{\mu}_1, \boldsymbol{\mu}_2 \in \mathbb{R}^d$ as $\boldsymbol{\mu}_1 = \mathbf{0}$ and $\boldsymbol{\mu}_2 = (1, 1, \ldots, 1, 0, 0, \ldots, 0)^\top$, where the number of 1's is 10. It is easy to get that $\boldsymbol{\beta}^*$ is a sparse vector with 11 nonzero entries. We set $r = 0.5$, which means that there are equal number of samples from the two normal distributions on each machine.

We use the following metrics to evaluate the performance of algorithms for comparison: the $\ell_2$ and $\ell_\infty$ norms of parameter estimation error. Additionally, to measure the support recovery, $F_1$ score is used to measure the overlap of estimated supports and true supports. The definition of $F_1$ score is as follows

$$F_1 = \frac{2 \cdot \text{precision} \cdot \text{recall}}{(\text{precision} + \text{recall})},$$

where precision $= |\operatorname{supp}(\bar{\boldsymbol{\beta}}) \cap \operatorname{supp}(\boldsymbol{\beta}^*)| / |\operatorname{supp}(\bar{\boldsymbol{\beta}})|$ and recall $= |\operatorname{supp}(\bar{\boldsymbol{\beta}}) \cap \operatorname{supp}(\boldsymbol{\beta}^*)| / |\operatorname{supp}(\boldsymbol{\beta}^*)|$, where $\bar{\boldsymbol{\beta}}$ is some estimator. The $|\cdot|$ here means the cardinality of a set.

**Table 1:** The computation time of distributed LDA vs. centralized LDA ($m = 1$ indicates centralized algorithm).

| $m$ | 1 | 20 | 40 | 60 | 80 | 100 |
|---|---|---|---|---|---|---|
| time (in second) | 863.4 | 48.37 | 33.65 | 21.87 | 15.46 | 10.38 |

**Table 2:** Misclassification rates of different methods on the real dataset

| $m$ | Centralized SLDA | Naive Averaged SLDA | Distributed SLDA |
|---|---|---|---|
| 4 | $0.208 \pm 0.012$ | $0.329 \pm 0.035$ | $0.220 \pm 0.017$ |

For the centralized estimator and the naively averaged estimator, there is one regularization parameter $\lambda$. By the theoretical result, a proper choice of $\lambda$ should be in the order of $O(\sqrt{N^{-1} \log d})$ for centralized estimator, and $O(\sqrt{n^{-1} \log d})$ for naively averaged estimator. Therefore, we set $\lambda = C\sqrt{N^{-1} \log d}$ (or $C\sqrt{n^{-1} \log d}$) and tune $C$ by grid search. For the proposed estimator, other than $\lambda$, there are two more parameters to be tuned: $\lambda'$ and $t$. The theoretical result reveals that $\lambda'$ should be in the order of $O(\sqrt{n^{-1} \log d})$. Thus, we simply set $\lambda' = \lambda$. The parameter $t$ is tuned in a similar way as the tuning of $\lambda$. We report the best results for all methods for the sake of fairness.

To investigate the effect of number of machines $m$, we fix the total sample size $N = 10000$ and vary the number of machines. Figure 1 shows how the $F_1$ score and estimation error (in $\ell_2$ and $\ell_\infty$ norm) of the proposed estimator change as the number of machine grows. The widths of the curves represent the standard deviations of metrics such as $F_1$ scores and $\ell_2, \ell_\infty$ norms. The standard deviations are obtained after repeating the experiments 20 times.

From Figure 1, it can be seen that the proposed distributed LDA algorithm is comparable to the centralized LDA estimator in both support recovery and parameter estimation when $m$ is small, while the naive averaged estimator is much worse. Moreover, we can see that the estimation error of distributed LDA will be larger than that of centralized LDA as $m$ surpasses a certain threshold. This is consistent with the result of Theorem 4.6. That is, if $m$ is too big, the dominating term in the estimation error bound will be the second term, which depends on $m$.

Next we focus on the effect of averaging, we increase the number of machines $m$ linearly as the total sample size $N$, that is, the sample size on each machine $n$ is fixed. More specifically, we choose $n = 200$. Figure 2 displays the $F_1$ score, estimation error of our estimator, naively averaged estimator and centralized estimator in terms of $\ell_2$ and $\ell_\infty$ norms. We can see that the performance of distributed LDA is comparable to that of centralized LDA, while the performance of naively averaged estimator is much worse. We can also observe that as $N$ grows linearly with respect to $m$ (i.e., $n$ is fixed), the estimation error of distributed LDA decreases slower than that of centralized LDA. This is consistent with what Theorem 4.6 suggests: in (4.2) and (4.3), if $n$ is fixed and $m$ is growing, the first term of the error bounds will decrease because it is of the order $O(1/\sqrt{N})$. However, the second term in the error bounds will not decrease because it depends on $m/N = 1/n$. Therefore, the total estimation error of our algorithm will converge to a positive constant rather than zero.

The empirical computation time of distributed LDA and centralized LDA are summarized in Table 1. We set $d = 200, N = 10^6$ and vary $m$ between 20 and 100. For distributed LDA algorithm, we only take into account the time used in one local machine, rather than the total CPU time consumed by all machines, because the local computations are carried out in parallel. The experiment platform is Linux operating system with 2.8GHz CPU. From Table 1 we can see that the distributed algorithm has lower time cost than the centralized algorithm. Furthermore, Table 1 also demonstrates a near linear speedup with the number of machines, which is consistent with the time complexity analysis in Section 3.

## 5.2 Real Date Experiments

To verify the effectiveness of the proposed algorithm on real datasets, we use the Heart Disease dataset[1] to conduct the experiment. This dataset contains information of 920 heart disease patients across 4 hospitals. For each patient, there are 13 attributes associated, including gender, age, laboratory test results, etc. Every patient is labeled with the diagnosis result, i.e., whether he or she is diagnosed as heart disease. In the preprocessing step, we extend all categorical attributes into binary dummy variables. For the missing values in any numeric attributes in the dataset, we replace them with the average value of the attribute that it belongs to. After the preprocessing, we get 920 entries, each with 22 numerical attributes.

The dataset is naturally divided into 4 parts by the hospital where each patient is diagnosed. We consider each part as the local data stored in one machine. In every part, we randomly choose half of the data as the training set and the remaining half as the test set. To get a proper choice of parameters, as in the synthetic data experiment, we set $\lambda = C\sqrt{N^{-1} \log d}$ (or $C\sqrt{n^{-1} \log d}$), $\lambda' = \lambda$ and use 5-fold cross validation on the training set to tune $C$ and $t$. After the training phase, we test the misclassification rate of classifiers obtained by different methods on the test set. The experiment is repeated 10 times (i.e., training and test set splitting) and the averaged misclassification rates along with their standard deviations are reported in Table 2. It can be seen that the proposed method greatly decreases the misclassification rate compared with the naive averaged estimator, and achieves a comparable performance with the centralized LDA estimator. This verifies the effectiveness of our algorithm on real data.

## 6 Conclusions and Future Work

We proposed a communication efficient distributed algorithm for sparse linear discriminant analysis in the high dimensional regime. The key idea is constructing a local debiased estimator on each machine and averaging them over all machines. We addressed an important question that how to choose the number of machines such that the aggregated estimator will attain the same convergence rate as the centralized estimator. Experiments on both synthetic and real datasets corroborate our theory. In the future, we will extend our algorithm and theory to multi-class sparse LDA.

---

[1] https://archive.ics.uci.edu/ml/datasets/Heart+Disease

# A Proof of the Main Theory

Before we prove the main results, we first lay out a key lemma, which is crucial to establish the main theory.

**Lemma A.1.** *Under Assumptions 4.1, 4.2 and Condition 4.4, if $\lambda = C_1 K^2 \sqrt{\log d/(rn)} \|\boldsymbol{\beta}^*\|_1, \lambda' = C_2 K^2 M \sqrt{\log d/n}$, we have with probability at least $1 - 18m/d - 4/d$ that*

$$\left\| \frac{1}{m} \sum_{l=1}^{m} \widetilde{\boldsymbol{\beta}}^{(l)} - \boldsymbol{\beta}^* \right\|_\infty \leq C' M \sqrt{\frac{\log d}{N}} \|\boldsymbol{\beta}^*\|_1 + C'' \max(s, s') M \frac{m \log d}{N} \|\boldsymbol{\beta}^*\|_1,$$

*where $C'$ and $C''$ are constants.*

Lemma A.1 gives an upper bound on the $\ell_\infty$ estimation error for the averaged debiased estimator.

**Lemma A.2.** *Under Assumption 4.1, for any $l \in \{1, 2, \ldots, d\}$, if we define*

$$\widetilde{\boldsymbol{\Sigma}}^{(l)} = \frac{1}{n} \left[ \sum_{i=1}^{n_1} (\boldsymbol{X}_i^{(l)} - \boldsymbol{\mu}_1)(\boldsymbol{X}_i^{(l)} - \boldsymbol{\mu}_1)^\top + \sum_{i=1}^{n_2} (\boldsymbol{Y}_i^{(l)} - \boldsymbol{\mu}_2)(\boldsymbol{Y}_i^{(l)} - \boldsymbol{\mu}_2)^\top \right], \tag{A.1}$$

*then with probability at least $1 - 2/d$ the following inequality holds:*

$$\|\widehat{\boldsymbol{\Sigma}}^{(l)} - \widetilde{\boldsymbol{\Sigma}}^{(l)}\|_{\infty,\infty} \leq \frac{CK^2 \log d}{rn}.$$

Now we are ready to prove the main theorems and corollaries.

## A.1 Proof of Proposition 4.5

*Proof.* We denote $s^* = \max(s, s')$. For any index set $S$ satisfying $|S| \leq s^*$, and for any vector $\mathbf{u}$ in the cone $\{\mathbf{u} : \|\mathbf{u}_{S^c}\|_1 \leq \|\mathbf{u}_S\|_1\}$, we have

$$\mathbf{u}^\top \widehat{\boldsymbol{\Sigma}}^{(l)} \mathbf{u} = \mathbf{u}^\top \widetilde{\boldsymbol{\Sigma}}^{(l)} \mathbf{u} + \mathbf{u}^\top (\widehat{\boldsymbol{\Sigma}}^{(l)} - \widetilde{\boldsymbol{\Sigma}}^{(l)}) \mathbf{u} \geq \mathbf{u}^\top \widetilde{\boldsymbol{\Sigma}}^{(l)} \mathbf{u} - |\mathbf{u}^\top (\widehat{\boldsymbol{\Sigma}}^{(l)} - \widetilde{\boldsymbol{\Sigma}}^{(l)}) \mathbf{u}|. \tag{A.2}$$

Note that $\boldsymbol{\Sigma}^*$ satisfies the RE condition with parameter $(s^*, 1, \lambda_{\min}(\boldsymbol{\Sigma}^*))$ since $\boldsymbol{\Sigma}^*$ is a positive definite matrix following Assumption 4.1. From the definition of $\widetilde{\boldsymbol{\Sigma}}^{(l)}$ in (A.1) and Theorem E.7, it gives rise that there exist three universal constants $C_1$, $C_2$ and $C_3$, such that if $n$ satisfies the following inequality:

$$n > \frac{C_1 \rho^2(\boldsymbol{\Sigma}^*)}{\lambda_{\min}^2(\boldsymbol{\Sigma}^*)} s^* \log d,$$

then with probability at least $1 - C_2 \exp(-C_3 n)$, $\widetilde{\boldsymbol{\Sigma}}^{(l)}$ satisfies the RE condition with parameters $(s^*, 1, \lambda_{\min}(\boldsymbol{\Sigma}^*)/8)$, i.e., $\mathbf{u}^\top \widetilde{\boldsymbol{\Sigma}}^{(l)} \mathbf{u} \geq \lambda_{\min}(\boldsymbol{\Sigma}^*)/8 \|\mathbf{u}\|_2^2$. From Assumption 4.1 we get that $\lambda_{\min}(\boldsymbol{\Sigma}^*) \geq 1/K$ and $\rho^2(\boldsymbol{\Sigma}^*) \leq \lambda_{\max}(\boldsymbol{\Sigma}^*) \leq K$. Therefore we have

$$\frac{C_1 \rho^2(\boldsymbol{\Sigma}^*)}{\lambda_{\min}^2(\boldsymbol{\Sigma}^*)} s^* \log d \leq C_1 K^3 s^* \log d.$$

Next we give an bound on $|\mathbf{u}^\top(\widehat{\mathbf{\Sigma}}^{(l)} - \widetilde{\mathbf{\Sigma}}^{(l)})\mathbf{u}|$: we have

$$|\mathbf{u}^\top(\widehat{\mathbf{\Sigma}}^{(l)} - \widetilde{\mathbf{\Sigma}}^{(l)})\mathbf{u}| \leq \|\mathbf{u}\|_1 \cdot \|(\widehat{\mathbf{\Sigma}}^{(l)} - \widetilde{\mathbf{\Sigma}}^{(l)})\mathbf{u}\|_\infty \leq \|\mathbf{u}\|_1^2 \cdot \|\widehat{\mathbf{\Sigma}}^{(l)} - \widetilde{\mathbf{\Sigma}}^{(l)}\|_{\infty,\infty}, \tag{A.3}$$

where the first and second inequality follow from Hölder's inequality. Moreover, by $\|\mathbf{u}_{S^c}\|_1 \leq \|\mathbf{u}_S\|_1$ we have $\|\mathbf{u}\|_1 = \|\mathbf{u}_{S^c}\|_1 + \|\mathbf{u}_S\|_1 \leq 2\|\mathbf{u}_S\|_1$. Substituting it into (A.3) gives rise to

$$|\mathbf{u}^\top(\widehat{\mathbf{\Sigma}}^{(l)} - \widetilde{\mathbf{\Sigma}}^{(l)})\mathbf{u}| \leq (2\|\mathbf{u}_S\|_1)^2 \|\widehat{\mathbf{\Sigma}}^{(l)} - \widetilde{\mathbf{\Sigma}}^{(l)}\|_{\infty,\infty} \leq 4s^* \|\mathbf{u}_S\|_2^2 \cdot \|\widehat{\mathbf{\Sigma}}^{(l)} - \widetilde{\mathbf{\Sigma}}^{(l)}\|_{\infty,\infty}$$
$$\leq 4s^* \|\mathbf{u}\|_2^2 \cdot \|\widehat{\mathbf{\Sigma}}^{(l)} - \widetilde{\mathbf{\Sigma}}^{(l)}\|_{\infty,\infty},$$

where the second inequality follows from Cauchy-Schwartz inequality. By (B.10) with probability at least $1 - 2/d$ we have

$$|\mathbf{u}^\top(\widehat{\mathbf{\Sigma}}^{(l)} - \widetilde{\mathbf{\Sigma}}^{(l)})\mathbf{u}| \leq \frac{4s^* C_4 K^2 \log d}{rn}\|\mathbf{u}\|_2^2.$$

Applying this bound on (A.2) gives that

$$\mathbf{u}^\top \widehat{\mathbf{\Sigma}}^{(l)} \mathbf{u} \geq \left(\frac{\lambda_{\min}(\mathbf{\Sigma}^*)}{8} - \frac{4s^* C_4 K^2 \log d}{rn}\right)\|\mathbf{u}\|_2^2$$

with probability at least $1 - C_1 \exp(-C_2 n) - 2/d$. If we set $n > s^* r^{-1} \max\{64 C_4, C_1\} K^3 \log d$, it yields that $\mathbf{u}^\top \widehat{\mathbf{\Sigma}}^{(l)} \mathbf{u} \geq \lambda_{\min}(\mathbf{\Sigma}^*)/16$, i.e., $\widehat{\mathbf{\Sigma}}^{(l)}$ satisfies the RE condition with parameters $(s^*, 1, \lambda_{\min}(\mathbf{\Sigma}^*)/16)$. Applying union bound over $l \in \{1, 2, \ldots, d\}$, the probability of the RE condition to be satisfied by all $\widehat{\mathbf{\Sigma}}^{(l)}$'s is $1 - mC_1 \exp(-C_2 n) - 2m/d$. This completes the proof. □

### A.2 Proof of Theorem 4.6

*Proof.* From the definition of $\mathrm{HT}(\cdot, t)$ we have that $\forall \mathbf{u} \in \mathbb{R}^d, \forall j \in \{1, 2, \ldots, d\}, |(\mathrm{HT}(\mathbf{u}, t))_j - u_j| \leq t$. Hence $\|\mathrm{HT}(\mathbf{u}, t) - \mathbf{u}\|_\infty \leq t$. By triangle inequality, we have

$$\|\bar{\boldsymbol{\beta}} - \boldsymbol{\beta}^*\|_\infty \leq \left\|\bar{\boldsymbol{\beta}} - \frac{1}{m}\sum_{l=1}^m \widetilde{\boldsymbol{\beta}}^{(l)}\right\|_\infty + \left\|\frac{1}{m}\sum_{l=1}^m \widetilde{\boldsymbol{\beta}}^{(l)} - \boldsymbol{\beta}^*\right\|_\infty$$
$$\leq t + \left\|\frac{1}{m}\sum_{l=1}^m \widetilde{\boldsymbol{\beta}}^{(l)} - \boldsymbol{\beta}^*\right\|_\infty$$
$$\leq 2t = 2C'M\sqrt{\frac{\log d}{N}}\|\boldsymbol{\beta}^*\|_1 + 2C'' \max(s, s')M\frac{m \log d}{N}\|\boldsymbol{\beta}^*\|_1, \tag{A.4}$$

where the third inequality follows from Lemma A.1. Now we consider the error bound of the $\ell_2$ and $\ell_1$ norm. If the event $\mathcal{E} := \{\|\left(\sum_{l=1}^m \widetilde{\boldsymbol{\beta}}^{(l)}\right)/m - \boldsymbol{\beta}^*\|_\infty \leq t\}$ happens, for any $j \in S^c$, we have

$$\left|\frac{1}{m}\sum_{l=1}^m \widetilde{\beta}_j^{(l)}\right| = \left|\frac{1}{m}\sum_{l=1}^m \widetilde{\beta}_j^{(l)} - \beta_j^*\right| \leq t,$$

where the first equality follows from the fact that $\beta_j^* = 0$. By the truncation rule, we have $\bar{\beta}_j = 0$. Hence $\operatorname{supp}(\bar{\boldsymbol{\beta}}) \subseteq S$. Therefore, under event $\mathcal{E}$, we have

$$\|\bar{\boldsymbol{\beta}} - \boldsymbol{\beta}^*\|_2 = \|(\bar{\boldsymbol{\beta}} - \boldsymbol{\beta}^*)_S\|_2 \leq \sqrt{s}\|(\bar{\boldsymbol{\beta}} - \boldsymbol{\beta}^*)_S\|_\infty$$
$$\leq \sqrt{s}\|\bar{\boldsymbol{\beta}} - \boldsymbol{\beta}^*\|_\infty \leq 2\sqrt{s}t$$
$$= 2\sqrt{s}C'M\sqrt{\frac{\log d}{N}}\|\boldsymbol{\beta}^*\|_1 + 2\sqrt{s}C''\max(s,s')M\frac{m\log d}{N}\|\boldsymbol{\beta}^*\|_1,$$

where the second inequality follows from Cauchy-Schwartz inequality and the fourth inequality follows from (A.4). Similarly, we have

$$\|\bar{\boldsymbol{\beta}} - \boldsymbol{\beta}^*\|_1 = \|(\bar{\boldsymbol{\beta}} - \boldsymbol{\beta}^*)_S\|_1 \leq s\|(\bar{\boldsymbol{\beta}} - \boldsymbol{\beta}^*)_S\|_\infty \leq 2st$$
$$\leq 2sC'M\sqrt{\frac{\log d}{N}}\|\boldsymbol{\beta}^*\|_1 + 2sC''\max(s,s')M\frac{m\log d}{N}\|\boldsymbol{\beta}^*\|_1.$$

This completes the proof. □

## A.3 Proof of Corollary 4.8

*Proof.* Substituting (4.5) into (4.2), we obtain with probability at least $1 - 18m/d - 4/d$ that

$$\|\bar{\boldsymbol{\beta}} - \boldsymbol{\beta}^*\|_\infty \leq 2C'M\sqrt{\frac{\log d}{N}}\|\boldsymbol{\beta}^*\|_1 + 2C''\max(s,s')M\left(\frac{C'''}{\max(s,s')}\sqrt{\frac{N}{\log d}}\right)\frac{\log d}{N}\|\boldsymbol{\beta}^*\|_1$$
$$= CM\sqrt{\frac{\log d}{N}}\|\boldsymbol{\beta}^*\|_1.$$

Substituting (4.5) into (4.3), we obtain with probability at least $1 - 18m/d - 4/d$ that

$$\|\bar{\boldsymbol{\beta}} - \boldsymbol{\beta}^*\|_2 \leq 2\sqrt{s}C'M\sqrt{\frac{\log d}{N}}\|\boldsymbol{\beta}^*\|_1 + 2\sqrt{s}C''\max(s,s')M\left(\frac{C'''}{\max(s,s')}\sqrt{\frac{N}{\log d}}\right)\frac{\log d}{N}\|\boldsymbol{\beta}^*\|_1$$
$$= \sqrt{s}CM\sqrt{\frac{\log d}{N}}\|\boldsymbol{\beta}^*\|_1,$$

Substituting (4.5) into (4.4), we obtain with probability at least $1 - 18m/d - 4/d$ that

$$\|\bar{\boldsymbol{\beta}} - \boldsymbol{\beta}^*\|_1 \leq 2sC'M\sqrt{\frac{\log d}{N}}\|\boldsymbol{\beta}^*\|_1 + 2sC''\max(s,s')M\left(\frac{C'''}{\max(s,s')}\sqrt{\frac{N}{\log d}}\right)\frac{\log d}{N}\|\boldsymbol{\beta}^*\|_1$$
$$= sCM\sqrt{\frac{\log d}{N}}\|\boldsymbol{\beta}^*\|_1.$$

This completes the proof. □

## A.4 Proof of Theorem 4.10

*Proof.* We define $t$ as in (4.1), and the event $\mathcal{E}$ is defined as

$$\mathcal{E} := \left\{\left\|\frac{1}{m}\sum_{l=1}^m \widetilde{\boldsymbol{\beta}}^{(l)} - \boldsymbol{\beta}^*\right\|_\infty \leq t\right\}.$$

This event is the event that the conclusion of Lemma A.1 holds. From Lemma A.1 we know that $\mathcal{E}$ happens with probability at least $1 - 18m/d - 4/d$. Under the condition that $\beta_{\min} > 2t$, we have that

1. if $\beta_j^* > 0$, which means that $\beta_j^* > 2t$, it holds that $\bar{\beta}_j \geq \beta_j^* - |\bar{\beta}_j - \beta_j^*| > 2t - 2t = 0$.
2. if $\beta_j^* < 0$, which means that $\beta_j^* < -2t$, it holds that $\bar{\beta}_j \leq \beta_j^* + |\bar{\beta}_j - \beta_j^*| < -2t + 2t = 0$.
3. if $\beta_j^* = 0$, because event $\mathcal{E}$ happens we have $|(\sum_{l=1}^m \widetilde{\beta}_j^{(l)})/m| \leq t$. By the definition of the hard thresholding operator $\mathrm{HT}(\cdot, t)$, we have $\bar{\beta}_j = 0$.

Conclusively we have $\mathrm{sign}(\beta_j^*) = \mathrm{sign}(\bar{\beta}_j)$ for all $j \in \{1, 2, \ldots, d\}$ if the event $\mathcal{E}$ happens, which has a probability at least $1 - 18m/d - 4/d$. □

### A.5 Proof of Corollary 4.11

*Proof.* Substituting (4.7) into (4.6), we obtain that the condition of $\beta_{\min}$ in Theorem 4.10 can be rewritten as

$$\beta_{\min} > 2C'M\sqrt{\frac{\log d}{N}}\|\boldsymbol{\beta}^*\|_1 + 2C''\max(s, s')M\left(\frac{C_1}{\max(s, s')}\sqrt{\frac{N}{\log d}}\right)\frac{\log d}{N}\|\boldsymbol{\beta}^*\|_1$$

$$= CM\sqrt{\frac{\log d}{N}}\|\boldsymbol{\beta}^*\|_1.$$

It is obvious that our assumption on $\beta_{\min}$ satisfies this condition. Therefore by Theorem 4.10 we have with probability at least $1 - 18m/d - 4/d$ that $\mathrm{sign}(\beta_j^*) = \mathrm{sign}(\bar{\beta}_j)$ for all $j \in \{1, 2, \ldots, d\}$. □

## B Proof of Lemmas in Appendix A

First, we lay out some lemmas which are crucial to the proof of Lemma A.1.

**Lemma B.1.** *Let $\boldsymbol{X}_1, \boldsymbol{X}_2, \ldots, \boldsymbol{X}_n \in \mathbb{R}^d$ be i.i.d. random vectors following normal distribution $N(\boldsymbol{\mu}, \boldsymbol{\Sigma})$. And the sample mean $\bar{\boldsymbol{X}} = (\sum_{i=1}^n \boldsymbol{X}_i)/n$. Then the difference between $\bar{\boldsymbol{X}}$ and $\boldsymbol{\mu}$ can be bounded by*

$$\|\bar{\boldsymbol{X}} - \boldsymbol{\mu}\|_\infty \leq CK_{\boldsymbol{X}}\sqrt{\frac{\log d}{n}}$$

*with probability at least $1 - 1/d$, where $C$ is an absolute constant, $K_{\boldsymbol{X}} = \|\boldsymbol{X}_1\|_{\psi_2}$.*

**Lemma B.2.** *Let $\boldsymbol{X}_1, \boldsymbol{X}_2, \ldots, \boldsymbol{X}_n$ be i.i.d. random vectors following multivariate normal distribution with zero mean and covariance matrix $\boldsymbol{\Sigma} \in \mathbb{R}^{d \times d}$, then with probability at least $1 - 2/d$, the following inequality holds:*

$$\left\|\frac{1}{n}\sum_{i=1}^n \boldsymbol{X}_i\boldsymbol{X}_i^\top - \boldsymbol{\Sigma}\right\|_{\infty,\infty} \leq CK_{\boldsymbol{X}}^2\sqrt{\frac{\log d}{n}},$$

*where $C$ is an absolute constant, and $K_{\boldsymbol{X}} = \|\boldsymbol{X}_1\|_{\psi_2}$.*

**Lemma B.3.** *Under Assumption 4.1, for any $l \in \{1, 2, \ldots, m\}$, we have with probability at least $1 - 4/d$ that*

$$\left\|\widehat{\boldsymbol{\Sigma}}^{(l)} - \boldsymbol{\Sigma}^*\right\|_{\infty,\infty} \leq C_2 K^2 \sqrt{\frac{\log d}{n}}, \tag{B.1}$$

*where $C_1$ and $C_2$ are absolute constants.*

**Lemma B.4.** *Under Assumption 4.1 and Condition 4.4, if we set $\lambda$ as*

$$\lambda \geq CK^2 \sqrt{\frac{\log d}{rn}} \|\boldsymbol{\beta}^*\|_1, \tag{B.2}$$

*then we have with probability at least $1 - 6/d$ that*

$$\|\widehat{\boldsymbol{\beta}}^{(l)} - \boldsymbol{\beta}^*\|_1 \leq \frac{128\lambda s}{\lambda_{\min}(\boldsymbol{\Sigma}^*)}. \tag{B.3}$$

**Lemma B.5.** *Under Assumption 4.1, 4.2 and Condition 4.4, if we set $\lambda'$*

$$\lambda' \geq CK^2 M \sqrt{\frac{\log d}{n}}, \tag{B.4}$$

*then for each $l$ with probability at least $1 - 4/d$ we have*

$$\|\widehat{\boldsymbol{\theta}}_j^{(l)} - \boldsymbol{\theta}^*\|_1 \leq \frac{128\lambda' s'}{\lambda_{\min}(\boldsymbol{\Sigma}^*)}$$

*for all $j \in \{1, 2, \ldots, d\}$.*

Note that Lemma B.5 implies that for each machine,

$$\|\widehat{\boldsymbol{\Theta}}^{(l)} - \boldsymbol{\Theta}^*\|_1 \leq \frac{128\lambda' s'}{\lambda_{\min}(\boldsymbol{\Sigma}^*)}$$

holds with probability at least $1 - 4/d$.

## B.1 Proof of Lemma A.1

*Proof.* Substituting the definition of $\widetilde{\boldsymbol{\beta}}^{(l)}$, we have

$$\left\|\frac{1}{m}\sum_{l=1}^m \widetilde{\boldsymbol{\beta}}^{(l)} - \boldsymbol{\beta}^*\right\|_\infty = \left\|\frac{1}{m}\sum_{l=1}^m \widehat{\boldsymbol{\beta}}^{(l)} - \boldsymbol{\beta}^* - \widehat{\boldsymbol{\Theta}}^{(l)\top}(\widehat{\boldsymbol{\Sigma}}^{(l)}\widehat{\boldsymbol{\beta}}^{(l)} - \widehat{\boldsymbol{\mu}}_d^{(l)})\right\|_\infty$$

$$= \left\|\frac{1}{m}\sum_{l=1}^m \big(\mathbf{I} - \widehat{\boldsymbol{\Theta}}^{(l)\top}\widehat{\boldsymbol{\Sigma}}^{(l)}\big)\big(\widehat{\boldsymbol{\beta}}^{(l)} - \boldsymbol{\beta}^*\big) - \widehat{\boldsymbol{\Theta}}^{(l)\top}\widehat{\boldsymbol{\Sigma}}^{(l)}\boldsymbol{\beta}^* + \widehat{\boldsymbol{\Theta}}^{(l)\top}\widehat{\boldsymbol{\mu}}_d^{(l)}\right\|_\infty$$

$$= \left\|\frac{1}{m}\sum_{l=1}^m \big(\mathbf{I} - \widehat{\boldsymbol{\Theta}}^{(l)\top}\widehat{\boldsymbol{\Sigma}}^{(l)}\big)\big(\widehat{\boldsymbol{\beta}}^{(l)} - \boldsymbol{\beta}^*\big) + \widehat{\boldsymbol{\Theta}}^{(l)\top}\big(\widehat{\boldsymbol{\mu}}_d^{(l)} - \widehat{\boldsymbol{\Sigma}}^{(l)}\boldsymbol{\beta}^*\big)\right\|_\infty. \tag{B.5}$$

Using triangle inequality, we can split (B.5) into two terms:

$$\left\|\frac{1}{m}\sum_{l=1}^{m}\widetilde{\boldsymbol{\beta}}^{(l)} - \boldsymbol{\beta}^*\right\|_\infty \leq \underbrace{\left\|\frac{1}{m}\sum_{l=1}^{m}\left(\mathbf{I} - \widehat{\boldsymbol{\Theta}}^{(l)\top}\widehat{\boldsymbol{\Sigma}}^{(l)}\right)\left(\widehat{\boldsymbol{\beta}}^{(l)} - \boldsymbol{\beta}^*\right)\right\|_\infty}_{I_1} + \underbrace{\left\|\frac{1}{m}\sum_{l=1}^{m}\widehat{\boldsymbol{\Theta}}^{(l)\top}\left(\widehat{\boldsymbol{\mu}}_d^{(l)} - \widehat{\boldsymbol{\Sigma}}^{(l)}\boldsymbol{\beta}^*\right)\right\|_\infty}_{I_2}.$$

Firstly we give an upper bound of $I_1$. By triangle inequality we have

$$I_1 \leq \frac{1}{m}\sum_{l=1}^{m}\left\|\left(\mathbf{I} - \widehat{\boldsymbol{\Theta}}^{(l)\top}\widehat{\boldsymbol{\Sigma}}^{(l)}\right)\left(\widehat{\boldsymbol{\beta}}^{(l)} - \boldsymbol{\beta}^*\right)\right\|_\infty$$

$$\leq \frac{1}{m}\sum_{l=1}^{m}\left\|\mathbf{I} - \widehat{\boldsymbol{\Theta}}^{(l)\top}\widehat{\boldsymbol{\Sigma}}^{(l)}\right\|_{\infty,\infty} \cdot \left\|\widehat{\boldsymbol{\beta}}^{(l)} - \boldsymbol{\beta}^*\right\|_1.$$

By (3.2) and Lemma (B.4) we have with probability at least $1 - 6m/d$ that

$$I_1 \leq \frac{128\lambda'\lambda s}{\lambda_{\min}(\boldsymbol{\Sigma}^*)} = \frac{128sC_1C_2K^4M\log d}{\lambda_{\min}(\boldsymbol{\Sigma}^*)rn}\|\boldsymbol{\beta}^*\|_1 \leq \frac{sC_3M\log d}{n}\|\boldsymbol{\beta}^*\|_1,$$

where $C_3$ is a constant. Now we take a closer look at $I_2$. This term can be further rewritten as

$$I_2 = \left\|\frac{1}{m}\sum_{l=1}^{m}\boldsymbol{\Theta}^{*\top}\left(\widehat{\boldsymbol{\mu}}_d^{(l)} - \widehat{\boldsymbol{\Sigma}}^{(l)}\boldsymbol{\beta}^*\right) + (\widehat{\boldsymbol{\Theta}}^{(l)} - \widehat{\boldsymbol{\Theta}}^*)^\top\left(\widehat{\boldsymbol{\mu}}_d^{(l)} - \widehat{\boldsymbol{\Sigma}}^{(l)}\boldsymbol{\beta}^*\right)\right\|_\infty$$

$$\leq \underbrace{\left\|\frac{1}{m}\sum_{l=1}^{m}\boldsymbol{\Theta}^{*\top}\left(\widehat{\boldsymbol{\mu}}_d^{(l)} - \widehat{\boldsymbol{\Sigma}}^{(l)}\boldsymbol{\beta}^*\right)\right\|_\infty}_{I_3} + \underbrace{\left\|\frac{1}{m}\sum_{l=1}^{m}(\widehat{\boldsymbol{\Theta}}^{(l)} - \widehat{\boldsymbol{\Theta}}^*)^\top\left(\widehat{\boldsymbol{\mu}}_d^{(l)} - \widehat{\boldsymbol{\Sigma}}^{(l)}\boldsymbol{\beta}^*\right)\right\|_\infty}_{I_4}.$$

Firstly we bound $I_4$. By triangle inequality we have

$$I_4 \leq \frac{1}{m}\sum_{l=1}^{m}\left\|(\widehat{\boldsymbol{\Theta}}^{(l)} - \widehat{\boldsymbol{\Theta}}^*)^\top\left(\widehat{\boldsymbol{\mu}}_d^{(l)} - \widehat{\boldsymbol{\Sigma}}^{(l)}\boldsymbol{\beta}^*\right)\right\|_\infty \leq \frac{1}{m}\sum_{l=1}^{m}\left\|(\widehat{\boldsymbol{\Theta}}^{(l)} - \widehat{\boldsymbol{\Theta}}^*)^\top\right\|_\infty \cdot \left\|\widehat{\boldsymbol{\mu}}_d^{(l)} - \widehat{\boldsymbol{\Sigma}}^{(l)}\boldsymbol{\beta}^*\right\|_\infty.$$

In the proof of Lemma B.4, we get the conclusion that $\|\widehat{\boldsymbol{\mu}}_d^{(l)} - \widehat{\boldsymbol{\Sigma}}^{(l)}\boldsymbol{\beta}^*\|_\infty \leq \lambda$ with probability at least $1 - 6/d$. Moreover, using Lemma B.5 and union bound, we have with probability at least $1 - 10m/d$ that

$$I_4 \leq \frac{128\lambda\lambda's'}{\lambda_{\min}(\boldsymbol{\Sigma}^*)} = \frac{128s'C_1C_2K^4M\log d}{\lambda_{\min}(\boldsymbol{\Sigma}^*)rn}\|\boldsymbol{\beta}^*\|_1 \leq \frac{s'C_4M\log d}{n}\|\boldsymbol{\beta}^*\|_1,$$

where $C_4$ is a constant. Finally, we give a bound on $I_3$. This term can be rewritten as follows:

$$I_3 = \left\|\frac{1}{m}\sum_{l=1}^{m}\boldsymbol{\Theta}^{*\top}(\widehat{\boldsymbol{\mu}}_d^{(l)} - \boldsymbol{\mu}_d + \boldsymbol{\Sigma}^*\boldsymbol{\beta}^* - \widehat{\boldsymbol{\Sigma}}^{(l)}\boldsymbol{\beta}^*)\right\|_\infty$$

$$\leq \left\|\frac{1}{m}\sum_{l=1}^{m}\boldsymbol{\Theta}^{*\top}(\widehat{\boldsymbol{\mu}}_d^{(l)} - \boldsymbol{\mu}_d)\right\|_\infty + \left\|\frac{1}{m}\sum_{l=1}^{m}\boldsymbol{\Theta}^{*\top}(\boldsymbol{\Sigma}^* - \widehat{\boldsymbol{\Sigma}}^{(l)})\boldsymbol{\beta}^*\right\|_\infty$$

$$\leq \underbrace{\left\|\frac{1}{m}\sum_{l=1}^{m}\boldsymbol{\Theta}^{*\top}(\widehat{\boldsymbol{\mu}}_d^{(l)} - \boldsymbol{\mu}_d)\right\|_\infty}_{I_5} + \underbrace{\left\|\frac{1}{m}\sum_{l=1}^{m}\boldsymbol{\Theta}^{*\top}(\boldsymbol{\Sigma}^* - \widetilde{\boldsymbol{\Sigma}}^{(l)})\boldsymbol{\beta}^*\right\|_\infty}_{I_6} + \underbrace{\frac{1}{m}\sum_{l=1}^{m}\left\|\boldsymbol{\Theta}^{*\top}(\widetilde{\boldsymbol{\Sigma}}^{(l)} - \widehat{\boldsymbol{\Sigma}}^{(l)})\boldsymbol{\beta}^*\right\|_\infty}_{I_7},$$

where the first equality uses the fact that $\boldsymbol{\mu}_d = \boldsymbol{\Sigma}^*\boldsymbol{\beta}^*$. Furthermore, $I_7$ can be bounded by

$$I_7 \leq \frac{1}{m}\sum_{l=1}^m \|\boldsymbol{\Theta}^{*\top}\|_\infty \cdot \|(\widetilde{\boldsymbol{\Sigma}}^{(l)} - \widehat{\boldsymbol{\Sigma}}^{(l)})\boldsymbol{\beta}^*\|_\infty$$

$$\leq \frac{1}{m}\sum_{l=1}^m \|\boldsymbol{\Theta}^*\|_1 \cdot \|\widetilde{\boldsymbol{\Sigma}}^{(l)} - \widehat{\boldsymbol{\Sigma}}^{(l)}\|_{\infty,\infty} \cdot \|\boldsymbol{\beta}^*\|_1.$$

By Lemma B.3, we have with probability at least $1 - 2m/d$ that

$$I_7 \leq \frac{1}{m}\sum_{l=1}^m M \frac{C_5 \log d}{n} = \frac{C_5 M \log d}{n},$$

where the first inequality follows from the fact that $\|\boldsymbol{\Theta}^*\|_1 \leq M$. In terms of $I_5$, we have

$$I_5 = \left\|\boldsymbol{\Theta}^{*\top}\left[\left(\frac{1}{m}\sum_{l=1}^m \widehat{\boldsymbol{\mu}}_1^{(l)} - \boldsymbol{\mu}_1\right) - \left(\frac{1}{m}\sum_{l=1}^m \widehat{\boldsymbol{\mu}}_2^{(l)} - \boldsymbol{\mu}_2\right)\right]\right\|_\infty$$

$$\leq \|\boldsymbol{\Theta}^{*\top}\|_\infty \cdot (\|\widehat{\boldsymbol{\mu}}_1 - \boldsymbol{\mu}_1\|_\infty + \|\widehat{\boldsymbol{\mu}}_2 - \boldsymbol{\mu}_2\|_\infty),$$

where $\widehat{\boldsymbol{\mu}}_1 = (\sum_{l=1}^m \widehat{\boldsymbol{\mu}}_1^{(l)})/m, \widehat{\boldsymbol{\mu}}_2 = (\sum_{l=1}^m \widehat{\boldsymbol{\mu}}_2^{(l)})/m$. By Lemma B.1, we have with probability at least $1 - 1/d$ that

$$\|\widehat{\boldsymbol{\mu}}_1 - \boldsymbol{\mu}_1\|_\infty \leq C_6 K\sqrt{\frac{\log d}{mn_1}}.$$

Similarly, we have with probability at least $1 - 1/d$ that

$$\|\widehat{\boldsymbol{\mu}}_2 - \boldsymbol{\mu}_2\|_\infty \leq C_6 K\sqrt{\frac{\log d}{mn_2}}.$$

Therefore we have with probability at least $1 - 2/d$ that

$$I_5 \leq C_6 MK\left(\sqrt{\frac{\log d}{mn_1}} + \sqrt{\frac{\log d}{mn_2}}\right).$$

In terms of $I_6$, we can apply similar procedure. Let us denote $\widetilde{\boldsymbol{\Sigma}} = (\sum_{l=1}^m \widetilde{\boldsymbol{\Sigma}}^{(l)})/m$. By Lemma B.2 we have with probability at least $1 - 2/d$ that

$$\|\widetilde{\boldsymbol{\Sigma}} - \boldsymbol{\Sigma}^*\|_{\infty,\infty} \leq C_7 K^2 \sqrt{\frac{\log d}{N}}.$$

Therefore $I_6$ can be bounded by

$$I_6 = \|\boldsymbol{\Theta}^{*\top}(\boldsymbol{\Sigma}^* - \widetilde{\boldsymbol{\Sigma}})\boldsymbol{\beta}^*\|_\infty \leq \|\boldsymbol{\Theta}^{*\top}\|_\infty \cdot \|(\boldsymbol{\Sigma}^* - \widetilde{\boldsymbol{\Sigma}})\boldsymbol{\beta}^*\|_\infty \leq M \cdot \|\boldsymbol{\Sigma}^* - \widetilde{\boldsymbol{\Sigma}}\|_{\infty,\infty} \cdot \|\boldsymbol{\beta}^*\|_1$$

$$\leq C_7 MK^2 \sqrt{\frac{\log d}{N}} \|\boldsymbol{\beta}^*\|_1$$

with probability at least $1 - 2/d$. Combining the bound of $I_1, I_4, I_5, I_6$ and $I_7$, we get that

$$\left\| \frac{1}{m} \sum_{l=1}^m \widetilde{\boldsymbol{\beta}}^{(l)} - \boldsymbol{\beta}^* \right\|_\infty \leq \frac{sC_3 M \log d}{n} \|\boldsymbol{\beta}^*\|_1 + \frac{s'C_4 M \log d}{n} \|\boldsymbol{\beta}^*\|_1 + \frac{C_5 M K^2 \log d}{n} +$$

$$C_6 M K \left( \sqrt{\frac{\log d}{mn_1}} + \sqrt{\frac{\log d}{mn_2}} \right) + C_7 M K^2 \sqrt{\frac{\log d}{N}} \|\boldsymbol{\beta}^*\|_1$$

$$\leq C' M \sqrt{\frac{\log d}{N}} \|\boldsymbol{\beta}^*\|_1 + C'' \max(s, s') M \frac{m \log d}{N} \|\boldsymbol{\beta}^*\|_1$$

with probability at least $1 - 18m/d - 4/d$, where $C'$ and $C''$ are constants. $\square$

## B.2 Proof of Lemma A.2

*Proof.* By the definition of $\widehat{\boldsymbol{\Sigma}}^{(l)}$ and $\widetilde{\boldsymbol{\Sigma}}^{(l)}$, we have

$$\|\widehat{\boldsymbol{\Sigma}}^{(l)} - \widetilde{\boldsymbol{\Sigma}}^{(l)}\|_{\infty,\infty} = \frac{1}{n} \left\| \sum_{i=1}^{n_1} \left( (\boldsymbol{X}_i^{(l)} - \widehat{\boldsymbol{\mu}}_1^{(l)})(\boldsymbol{X}_i^{(l)} - \widehat{\boldsymbol{\mu}}_1^{(l)})^\top - (\boldsymbol{X}_i^{(l)} - \boldsymbol{\mu}_1)(\boldsymbol{X}_i^{(l)} - \boldsymbol{\mu}_1)^\top \right) \right.$$

$$\left. + \sum_{i=1}^{n_2} \left( (\boldsymbol{Y}_i^{(l)} - \widehat{\boldsymbol{\mu}}_2^{(l)})(\boldsymbol{Y}_i^{(l)} - \widehat{\boldsymbol{\mu}}_2^{(l)})^\top - (\boldsymbol{Y}_i^{(l)} - \boldsymbol{\mu}_2)(\boldsymbol{Y}_i^{(l)} - \boldsymbol{\mu}_2)^\top \right) \right\|_{\infty,\infty}. \quad (B.6)$$

Note that

$$\sum_{i=1}^{n_1} \left( (\boldsymbol{X}_i^{(l)} - \widehat{\boldsymbol{\mu}}_1^{(l)})(\boldsymbol{X}_i^{(l)} - \widehat{\boldsymbol{\mu}}_1^{(l)})^\top - (\boldsymbol{X}_i^{(l)} - \boldsymbol{\mu}_1)(\boldsymbol{X}_i^{(l)} - \boldsymbol{\mu}_1)^\top \right)$$

$$= \sum_{i=1}^{n_1} \boldsymbol{X}_i^{(l)} \boldsymbol{X}_i^{(l)\top} - n_1 \widehat{\boldsymbol{\mu}}_1^{(l)} \widehat{\boldsymbol{\mu}}_1^{(l)\top} - n_1 \widehat{\boldsymbol{\mu}}_1^{(l)} \widehat{\boldsymbol{\mu}}_1^{(l)\top} + n_1 \widehat{\boldsymbol{\mu}}_1^{(l)} \widehat{\boldsymbol{\mu}}_1^{(l)\top}$$

$$- \sum_{i=1}^{n_1} \boldsymbol{X}_i^{(l)} \boldsymbol{X}_i^{(l)\top} + n_1 \widehat{\boldsymbol{\mu}}_1^{(l)} \boldsymbol{\mu}_1^\top + n_1 \boldsymbol{\mu}_1 \widehat{\boldsymbol{\mu}}_1^{(l)\top} - n_1 \boldsymbol{\mu}_1 \boldsymbol{\mu}_1^\top$$

$$= -n_1 \widehat{\boldsymbol{\mu}}_1^{(l)} \widehat{\boldsymbol{\mu}}_1^{(l)\top} + n_1 \widehat{\boldsymbol{\mu}}_1^{(l)} \boldsymbol{\mu}_1^\top + n_1 \boldsymbol{\mu}_1 \widehat{\boldsymbol{\mu}}_1^{(l)\top} - n_1 \boldsymbol{\mu}_1 \boldsymbol{\mu}_1^\top$$

$$= -n_1 (\widehat{\boldsymbol{\mu}}_1^{(l)} - \boldsymbol{\mu}_1)(\widehat{\boldsymbol{\mu}}_1^{(l)} - \boldsymbol{\mu}_1)^\top. \quad (B.7)$$

Similarly, we have

$$\sum_{i=1}^{n_2} \left( (\boldsymbol{Y}_i^{(l)} - \widehat{\boldsymbol{\mu}}_2^{(l)})(\boldsymbol{Y}_i^{(l)} - \widehat{\boldsymbol{\mu}}_2^{(l)})^\top - (\boldsymbol{Y}_i^{(l)} - \boldsymbol{\mu}_2)(\boldsymbol{Y}_i^{(l)} - \boldsymbol{\mu}_2)^\top \right) = -n_2 (\widehat{\boldsymbol{\mu}}_2^{(l)} - \boldsymbol{\mu}_2)(\widehat{\boldsymbol{\mu}}_2^{(l)} - \boldsymbol{\mu}_2)^\top. \quad (B.8)$$

Substituting (B.7) and (B.8) into (B.6) and using the triangle inequality, we can obtain

$$\|\widehat{\boldsymbol{\Sigma}}^{(l)} - \widetilde{\boldsymbol{\Sigma}}^{(l)}\|_{\infty,\infty} \leq \frac{n_1}{n} \|(\widehat{\boldsymbol{\mu}}_1^{(l)} - \boldsymbol{\mu}_1)(\widehat{\boldsymbol{\mu}}_1^{(l)} - \boldsymbol{\mu}_1)^\top\|_{\infty,\infty} + \frac{n_2}{n} \|(\widehat{\boldsymbol{\mu}}_2^{(l)} - \boldsymbol{\mu}_2)(\widehat{\boldsymbol{\mu}}_2^{(l)} - \boldsymbol{\mu}_2)^\top\|_{\infty,\infty}$$

$$= \frac{n_1}{n} \|\widehat{\boldsymbol{\mu}}_1^{(l)} - \boldsymbol{\mu}_1\|_\infty^2 + \frac{n_2}{n} \|\widehat{\boldsymbol{\mu}}_2^{(l)} - \boldsymbol{\mu}_2\|_\infty^2. \quad (B.9)$$

By Lemma B.1, we know that $\|\widehat{\boldsymbol{\mu}}_1^{(l)} - \boldsymbol{\mu}_1\|_\infty \leq C'K\sqrt{\log d/n_1}$ with probability at least $1 - 1/d$ and $\|\widehat{\boldsymbol{\mu}}_2^{(l)} - \boldsymbol{\mu}_2\|_\infty \leq C'K\sqrt{\log d/n_2}$ with probability at least $1 - 1/d$. Submitting the two high probability bounds into (B.9) and using the union bound gives rise to

$$\|\widehat{\boldsymbol{\Sigma}}^{(l)} - \widetilde{\boldsymbol{\Sigma}}^{(l)}\|_{\infty,\infty} \leq \frac{C'^2 K^2 \log d}{\min(n_1, n_2)} \leq \frac{C'^2 K^2 \log d}{rn}. \tag{B.10}$$

This inequality holds with probability at least $1 - 2/d$. Setting $C = C'^2$ completes the proof. $\square$

## C Proof of Lemmas in Appendix B

First of all, we present some lemmas which are crucial to the proof of lemmas in this section.

**Lemma C.1.** *For the l-th machine, if the underlying true parameter $\boldsymbol{\beta}^*$ lies in the feasible set of the optimization problem (3.1), then the biased estimator $\widehat{\boldsymbol{\beta}}^{(l)}$ lies in the set $\{\boldsymbol{\beta} : \|(\boldsymbol{\beta} - \boldsymbol{\beta}^*)_{S^c}\|_1 \leq \|(\boldsymbol{\beta} - \boldsymbol{\beta}^*)_S\|_1\}$, where $S = \text{supp}(\boldsymbol{\beta}^*)$.*

**Lemma C.2.** *For the l-th machine, if the underlying true parameter $\boldsymbol{\theta}_j^*$ lies in the feasible set of the optimization problem (3.3), then the optimal solution $\widehat{\boldsymbol{\theta}}_j^{(l)}$ lies in the set $\{\boldsymbol{\theta} : \|(\boldsymbol{\theta} - \boldsymbol{\theta}_j^*)_{S_{\theta_j}^c}\|_1 \leq \|(\boldsymbol{\theta} - \boldsymbol{\theta}_j^*)_{S_{\theta_j}}\|_1\}$, where $S_{\theta_j} = \text{supp}(\boldsymbol{\theta}_j^*)$.*

### C.1 Proof of Lemma B.1

*Proof.* For the $j$-th component, by Theorem E.4, we have

$$\mathbb{P}(|\bar{X}_j - \mu_j| > t) \leq \exp\left(-\frac{C_1 n t^2}{K_{\boldsymbol{X}}^2}\right)$$

for any $t > 0$, where $C_1 > 0$ is an absolute constant. Using the union bound, we have

$$\mathbb{P}(\|\bar{\boldsymbol{X}} - \boldsymbol{\mu}\|_\infty > t) \leq d \exp\left(-\frac{C_1 n t^2}{K_{\boldsymbol{X}}^2}\right).$$

Taking $t = K_{\boldsymbol{X}}\sqrt{2\log d/C_1 n}$ gives that with probability at least $1 - 1/d$ the following inequality holds:

$$\|\bar{\boldsymbol{X}} - \boldsymbol{\mu}\|_\infty \leq K_{\boldsymbol{X}}\sqrt{\frac{2\log d}{C_1 n}}.$$

Setting $C = \sqrt{2/C_1}$ yields the conclusion of this lemma. $\square$

### C.2 Proof of Lemma B.2

*Proof.* Denote the $j$-th entry of $\boldsymbol{X}_i$ as $x_{ij}$. We have $(\boldsymbol{X}_i \boldsymbol{X}_i^\top)_{jk} = x_{ij}x_{ik}$ following sub-Exponential distribution. Because $\mathbb{E}(x_{ij}x_{ik}) = \Sigma_{jk}$ for all $i$, we know that $x_{ij}x_{ik} - \Sigma_{jk}$ is a centered sub-Exponential random variable. The $\psi_1$ norm of $x_{ij}x_{ik}$ can be bounded using Lemma E.6 as $\|x_{ij}x_{ik}\|_{\psi_1} \leq C_1 \max\{\|x_{ij}\|_{\psi_2}^2, \|x_{ik}\|_{\psi_2}^2\} \leq C_1 K_{\boldsymbol{X}}^2$. By Theorem E.5, for any $t > 0$ we have

$$\mathbb{P}\left(\left|\frac{1}{n}\sum_{i=1}^n x_{ij}x_{ik} - \Sigma_{jk}\right| \geq t\right) \leq 2\exp\left[-C_2 \min\left(\frac{t^2 n}{C_1^2 K_{\boldsymbol{X}}^4}, \frac{tn}{C_1 K_{\boldsymbol{X}}^2}\right)\right].$$

Using the union bound, we get the conclusion that

$$\mathbb{P}\bigg(\bigg\|\frac{1}{n}\sum_{i=1}^n \boldsymbol{X}_i\boldsymbol{X}_i^\top - \boldsymbol{\Sigma}\bigg\|_{\infty,\infty} \geq t\bigg) \leq 2d^2 \exp\bigg[-C_2 \min\bigg(\frac{t^2 n}{C_1^2 K_{\boldsymbol{X}}^4}, \frac{tn}{C_1 K_{\boldsymbol{X}}^2}\bigg)\bigg].$$

Setting $2d^2 \exp[-C_2 t^2 n/(C_1^2 K_{\boldsymbol{X}}^4)] = \delta$, we get

$$t = C_1 K_{\boldsymbol{X}}^2 \sqrt{\frac{\log(2d^2/\delta)}{C_2 n}}.$$

Setting $\delta = 2/d$ and $C = \sqrt{3}C_1/\sqrt{C_2}$, we get the conclusion that with probability at least $1 - 2/d$, the following inequality holds:

$$\bigg\|\frac{1}{n}\sum_{i=1}^n \boldsymbol{X}_i\boldsymbol{X}_i^\top - \boldsymbol{\Sigma}\bigg\|_{\infty,\infty} \leq CK_{\boldsymbol{X}}^2 \sqrt{\frac{\log d}{n}}.$$

□

## C.3 Proof of Lemma B.3

*Proof.* Following Neykov et al. (2015), using triangle inequality, we have

$$\big\|\widehat{\boldsymbol{\Sigma}}^{(l)} - \boldsymbol{\Sigma}^*\big\|_{\infty,\infty} \leq \big\|\widehat{\boldsymbol{\Sigma}}^{(l)} - \widetilde{\boldsymbol{\Sigma}}^{(l)}\big\|_{\infty,\infty} + \big\|\widetilde{\boldsymbol{\Sigma}}^{(l)} - \boldsymbol{\Sigma}^*\big\|_{\infty,\infty}.$$

Now we bound the first term. In Lemma A.2 we have got that with probability at least $1 - 2/d$, the first term is bounded by

$$\big\|\widehat{\boldsymbol{\Sigma}}^{(l)} - \widetilde{\boldsymbol{\Sigma}}^{(l)}\big\|_{\infty,\infty} \leq \frac{C'K^2 \log d}{rn}.$$

For the second term, note that in each machine, $\boldsymbol{X}_i^{(l)} - \boldsymbol{\mu}_1$'s and $\boldsymbol{Y}_i^{(l)} - \boldsymbol{\mu}_2$'s are i.i.d. random vectors following normal distribution with zero mean and covariance matrix $\boldsymbol{\Sigma}^*$. Hence by Lemma B.2 we have with probability at least $1 - 2/d$ that

$$\|\widetilde{\boldsymbol{\Sigma}}^{(l)} - \boldsymbol{\Sigma}^*\|_{\infty,\infty} \leq C''K^2 \sqrt{\frac{\log d}{n}}.$$

Combining the two high probability bounds together, we have with probability at least $1 - 4/d$ that

$$\|\widehat{\boldsymbol{\Sigma}}^{(l)} - \boldsymbol{\Sigma}^*\|_{\infty,\infty} \leq C''K^2 \sqrt{\frac{\log d}{n}} + \frac{C'K^2 \log d}{rn} \leq 2C''K^2 \sqrt{\frac{\log d}{n}}.$$

Setting $C_1 = C'^4/C''^2$ and $C_2 = 2C''$ completes the proof. □

## C.4 Proof of Lemma B.4

*Proof.* First we will show that with high probability the true parameter $\boldsymbol{\beta}^* = \boldsymbol{\Theta}^* \boldsymbol{\mu}_d$ satisfies the constraint in (3.1), i.e., with high probability the inequality $\|\widehat{\boldsymbol{\Sigma}}\boldsymbol{\beta}^* - \widehat{\boldsymbol{\mu}}_d^{(l)}\|_\infty < \lambda$ holds. To show this, we consider

$$
\begin{aligned}
\left\|\widehat{\boldsymbol{\Sigma}}^{(l)}\boldsymbol{\beta}^* - \widehat{\boldsymbol{\mu}}_d^{(l)}\right\|_\infty &= \left\|\boldsymbol{\Sigma}^*\boldsymbol{\beta}^* - \boldsymbol{\Sigma}^*\boldsymbol{\beta}^* + \widehat{\boldsymbol{\Sigma}}^{(l)}\boldsymbol{\beta}^* - \boldsymbol{\mu}_d + \boldsymbol{\mu}_d - \widehat{\boldsymbol{\mu}}_d^{(l)}\right\|_\infty \\
&\leq \|\boldsymbol{\Sigma}^*\boldsymbol{\beta}^* - \boldsymbol{\mu}_d\|_\infty + \|(\widehat{\boldsymbol{\Sigma}}^{(l)} - \boldsymbol{\Sigma}^*)\boldsymbol{\beta}^*\|_\infty + \|\boldsymbol{\mu}_1 - \boldsymbol{\mu}_2 - \widehat{\boldsymbol{\mu}}_1^{(l)} + \widehat{\boldsymbol{\mu}}_2^{(l)}\|_\infty \\
&\leq \|\boldsymbol{\Sigma}^*\boldsymbol{\beta}^* - \boldsymbol{\mu}_d\|_\infty + \|\widehat{\boldsymbol{\Sigma}}^{(l)} - \boldsymbol{\Sigma}^*\|_{\infty,\infty} \cdot \|\boldsymbol{\beta}^*\|_1 + \|\widehat{\boldsymbol{\mu}}_1^{(l)} - \boldsymbol{\mu}_1\|_\infty + \|\widehat{\boldsymbol{\mu}}_2^{(l)} - \boldsymbol{\mu}_2\|_\infty,
\end{aligned}
\tag{C.1}
$$

where the second inequality follows from triangle inequality and the definition of $\boldsymbol{\mu}_d$ and $\widehat{\boldsymbol{\mu}}_d^{(l)}$, and the third inequality follows from Hölder's inequality and triangle inequality. Note that by the definition of $\boldsymbol{\beta}^*$ we have $\boldsymbol{\Sigma}^*\boldsymbol{\beta}^* - \boldsymbol{\mu}_d = \boldsymbol{0}$. For other terms, by Lemma B.3 we have with probability at least $1 - 4/d$ that $\|\widehat{\boldsymbol{\Sigma}}^{(l)} - \boldsymbol{\Sigma}^*\|_{\infty,\infty} \leq C_1 K^2 \sqrt{\log d/n}$. Additionally Lemma B.1 gives that $\|\widehat{\boldsymbol{\mu}}_1^{(l)} - \boldsymbol{\mu}_1\|_\infty \leq C_2 K \sqrt{\log d/n_1}$ and $\|\widehat{\boldsymbol{\mu}}_2^{(l)} - \boldsymbol{\mu}_2\|_\infty \leq C_2 K \sqrt{\log d/n_2}$. Substituting the three high probability bounds into (C.1), we have with probability at least $1 - 6/d$ that

$$\|\widehat{\boldsymbol{\Sigma}}^{(l)}\boldsymbol{\beta}^* - \widehat{\boldsymbol{\mu}}_d^{(l)}\|_\infty \leq C_1 K^2 \sqrt{\frac{\log d}{n}} \|\boldsymbol{\beta}^*\|_1 + C_2 K \sqrt{\frac{\log d}{\min(n_1, n_2)}} \leq CK^2 \sqrt{\frac{\log d}{rn}} \|\boldsymbol{\beta}^*\|_1.$$

This means that if $\lambda$ satisfies (B.2), $\boldsymbol{\beta}^*$ will lie in the feasible set of (3.1) with probability at least $1 - 6/d$. Applying Lemma C.1 gives $\|(\widehat{\boldsymbol{\beta}}^{(l)} - \boldsymbol{\beta}^*)_{S^c}\|_1 \leq \|(\widehat{\boldsymbol{\beta}}^{(l)} - \boldsymbol{\beta}^*)_S\|_1$. By Condition 4.4 we have

$$(\widehat{\boldsymbol{\beta}}^{(l)} - \boldsymbol{\beta}^*)^\top \widehat{\boldsymbol{\Sigma}}^{(l)} (\widehat{\boldsymbol{\beta}}^{(l)} - \boldsymbol{\beta}^*) \geq \frac{\lambda_{\min}(\boldsymbol{\Sigma}^*)}{16} \|\widehat{\boldsymbol{\beta}}^{(l)} - \boldsymbol{\beta}^*\|_2^2 \geq \frac{\lambda_{\min}(\boldsymbol{\Sigma}^*)}{16} \|(\widehat{\boldsymbol{\beta}}^{(l)} - \boldsymbol{\beta}^*)_S\|_2^2. \tag{C.2}$$

Additionally, we have

$$\|\widehat{\boldsymbol{\Sigma}}^{(l)}(\widehat{\boldsymbol{\beta}}^{(l)} - \boldsymbol{\beta}^*)\|_\infty \leq \|\widehat{\boldsymbol{\Sigma}}^{(l)}\widehat{\boldsymbol{\beta}}^{(l)} - \widehat{\boldsymbol{\mu}}_d^{(l)}\|_\infty + \|\widehat{\boldsymbol{\Sigma}}^{(l)}\boldsymbol{\beta}^* - \widehat{\boldsymbol{\mu}}_d^{(l)}\|_\infty \leq 2\lambda, \tag{C.3}$$

where the second inequality follows from the fact that both $\widehat{\boldsymbol{\beta}}^{(l)}$ and $\boldsymbol{\beta}^*$ are feasible solutions of optimization problem (3.1). Therefore we have

$$
\begin{aligned}
(\widehat{\boldsymbol{\beta}}^{(l)} - \boldsymbol{\beta}^*)^\top \widehat{\boldsymbol{\Sigma}}^{(l)} (\widehat{\boldsymbol{\beta}}^{(l)} - \boldsymbol{\beta}^*) &\leq \|\widehat{\boldsymbol{\Sigma}}^{(l)}(\widehat{\boldsymbol{\beta}}^{(l)} - \boldsymbol{\beta}^*)\|_\infty \cdot \|\widehat{\boldsymbol{\beta}}^{(l)} - \boldsymbol{\beta}^*\|_1 \leq 2\lambda \|\widehat{\boldsymbol{\beta}}^{(l)} - \boldsymbol{\beta}^*\|_1 \\
&\leq 4\lambda \|(\widehat{\boldsymbol{\beta}}^{(l)} - \boldsymbol{\beta}^*)_S\|_1 \\
&\leq 4\lambda \sqrt{s} \|(\widehat{\boldsymbol{\beta}}^{(l)} - \boldsymbol{\beta}^*)_S\|_2,
\end{aligned}
\tag{C.4}
$$

where the first inequality follows from Hölder's inequality, the second inequality follows from (C.3), the third follows from the fact that $\|(\widehat{\boldsymbol{\beta}}^{(l)} - \boldsymbol{\beta}^*)_{S^c}\|_1 \leq \|(\widehat{\boldsymbol{\beta}}^{(l)} - \boldsymbol{\beta}^*)_S\|_1$ and the last follows from Cauchy-Schwartz inequality. Combining (C.4) and (C.2) gives that

$$\|(\widehat{\boldsymbol{\beta}}^{(l)} - \boldsymbol{\beta}^*)_S\|_2 \leq \frac{64\lambda\sqrt{s}}{\lambda_{\min}(\boldsymbol{\Sigma}^*)}.$$

Based on this result we can provide the estimation error bound of $\widehat{\boldsymbol{\beta}}^{(l)}$ in terms of $\ell_1$ norm:

$$\|\widehat{\boldsymbol{\beta}}^{(l)} - \boldsymbol{\beta}^*\|_1 \leq 2\|(\widehat{\boldsymbol{\beta}}^{(l)} - \boldsymbol{\beta}^*)_S\|_1 \leq 2\sqrt{s}\|(\widehat{\boldsymbol{\beta}}^{(l)} - \boldsymbol{\beta}^*)_S\|_2 \leq \frac{128\lambda s}{\lambda_{\min}(\boldsymbol{\Sigma}^*)}.$$

$\square$

## C.5 Proof of Lemma B.5

*Proof.* First we will show that with high probability the true parameter $\boldsymbol{\theta}_j^*$ satisfies the constraint in (3.3), i.e., with high probability the inequality $\|\widehat{\boldsymbol{\Sigma}}\boldsymbol{\theta}_j^* - \mathbf{e}_j\|_\infty < \lambda'$ holds. To show this, we consider

$$\left\|\widehat{\boldsymbol{\Sigma}}^{(l)}\boldsymbol{\theta}_j^* - \mathbf{e}_j\right\|_\infty = \left\|\widehat{\boldsymbol{\Sigma}}^{(l)}\boldsymbol{\theta}_j^* - \boldsymbol{\Sigma}^*\boldsymbol{\theta}_j^*\right\|_\infty \le \|\widehat{\boldsymbol{\Sigma}}^{(l)} - \boldsymbol{\Sigma}^*\|_{\infty,\infty} \cdot \|\boldsymbol{\theta}_j^*\|_1, \tag{C.5}$$

where the first equality follows from the fact that $\boldsymbol{\Sigma}^*\boldsymbol{\theta}_j^* = \mathbf{e}_j$, and second inequality follows from Hölder's inequality. By Lemma B.3 we have with probability at least $1 - 4/d$ that $\|\widehat{\boldsymbol{\Sigma}}^{(l)} - \boldsymbol{\Sigma}^*\|_{\infty,\infty} \le CK^2\sqrt{\log d/n}$. Assumption 4.2 indicates that $\|\boldsymbol{\theta}_j^*\|_1 \le M$ for all $j$. Substituting the two high probability bounds into (C.5), for all $j \in \{1, 2, \ldots, d\}$ we have with probability at least $1 - 4/d$ that

$$\|\widehat{\boldsymbol{\Sigma}}^{(l)}\boldsymbol{\theta}_j^* - \mathbf{e}_j\|_\infty \le CK^2 M \sqrt{\frac{\log d}{n}}.$$

This means that if $\lambda'$ satisfies (B.4), $\boldsymbol{\theta}_j^*$ will lie in the feasible set of (3.1) with probability at least $1 - 4/d$. Applying Lemma C.2 gives $\|(\widehat{\boldsymbol{\theta}}_j^{(l)} - \boldsymbol{\theta}_j^*)_{S_{\theta_j}^c}\|_1 \le \|(\widehat{\boldsymbol{\theta}}_j^{(l)} - \boldsymbol{\theta}_j^*)_{S_{\theta_j}}\|_1$. By Condition 4.4 we have

$$(\widehat{\boldsymbol{\theta}}_j^{(l)} - \boldsymbol{\theta}_j^*)^\top \widehat{\boldsymbol{\Sigma}}^{(l)} (\widehat{\boldsymbol{\theta}}_j^{(l)} - \boldsymbol{\theta}_j^*) \ge \frac{\lambda_{\min}(\boldsymbol{\Sigma}^*)}{16} \|\widehat{\boldsymbol{\theta}}_j^{(l)} - \boldsymbol{\theta}_j^*\|_2^2 \ge \frac{\lambda_{\min}(\boldsymbol{\Sigma}^*)}{16} \|(\widehat{\boldsymbol{\theta}}_j^{(l)} - \boldsymbol{\theta}_j^*)_{S_{\theta_j}}\|_2^2. \tag{C.6}$$

Additionally, we have

$$\|\widehat{\boldsymbol{\Sigma}}^{(l)}(\widehat{\boldsymbol{\theta}}_j^{(l)} - \boldsymbol{\theta}_j^*)\|_\infty \le \|\widehat{\boldsymbol{\Sigma}}^{(l)}\widehat{\boldsymbol{\theta}}_j^{(l)} - \mathbf{e}_j\|_\infty + \|\widehat{\boldsymbol{\Sigma}}^{(l)}\boldsymbol{\theta}_j^* - \mathbf{e}_j\|_\infty \le 2\lambda', \tag{C.7}$$

where the second inequality follows from the fact that both $\widehat{\boldsymbol{\theta}}_j^{(l)}$ and $\boldsymbol{\theta}_j^*$ are feasible solutions of optimization problem (3.3). Therefore we have

$$\begin{aligned}(\widehat{\boldsymbol{\theta}}_j^{(l)} - \boldsymbol{\theta}_j^*)^\top \widehat{\boldsymbol{\Sigma}}^{(l)} (\widehat{\boldsymbol{\theta}}_j^{(l)} - \boldsymbol{\theta}_j^*) &\le \|\widehat{\boldsymbol{\Sigma}}^{(l)}(\widehat{\boldsymbol{\theta}}_j^{(l)} - \boldsymbol{\theta}_j^*)\|_\infty \cdot \|\widehat{\boldsymbol{\theta}}_j^{(l)} - \boldsymbol{\theta}_j^*\|_1 \le 2\lambda'\|\widehat{\boldsymbol{\theta}}_j^{(l)} - \boldsymbol{\theta}_j^*\|_1 \\ &\le 4\lambda'\|(\widehat{\boldsymbol{\theta}}_j^{(l)} - \boldsymbol{\theta}_j^*)_{S_{\theta_j}}\|_1 \\ &\le 4\lambda'\sqrt{s'}\|(\widehat{\boldsymbol{\theta}}_j - \boldsymbol{\theta}_j^*)_{S_{\theta_j}}\|_2,\end{aligned} \tag{C.8}$$

where the first inequality follows from Hölder's inequality, the second inequality follows from (C.7), the third follows from the fact that $\|(\widehat{\boldsymbol{\theta}}_j^{(l)} - \boldsymbol{\theta}_j^*)_{S_{\theta_j}^c}\|_1 \le \|(\widehat{\boldsymbol{\theta}}_j^{(l)} - \boldsymbol{\theta}_j^*)_{S_{\theta_j}}\|_1$ and the last follows from Cauchy-Schwartz inequality. Combining (C.8) and (C.6) gives that

$$\|(\widehat{\boldsymbol{\theta}}_j - \boldsymbol{\theta}_j^*)_{S_{\theta_j}}\|_2 \le \frac{64\lambda'\sqrt{s'}}{\lambda_{\min}(\boldsymbol{\Sigma}^*)}.$$

Based on this result we can provide the estimation error bound of $\widehat{\boldsymbol{\theta}}_j^{(l)}$ in terms of $\ell_1$-norm:

$$\|\widehat{\boldsymbol{\theta}}_j^{(l)} - \boldsymbol{\theta}_j^*\|_1 \le 2\|(\widehat{\boldsymbol{\theta}}_j^{(l)} - \boldsymbol{\theta}_j^*)_{S_{\theta_j}}\|_1 \le 2\sqrt{s'}\|(\widehat{\boldsymbol{\theta}}_j^{(l)} - \boldsymbol{\theta}_j^*)_{S_{\theta_j}}\|_2 \le \frac{128\lambda' s'}{\lambda_{\min}(\boldsymbol{\Sigma}^*)}.$$

$\square$

# D   Proof of Auxilliary Lemmas in Appendix C

## D.1   Proof of Lemma C.1

*Proof.* In the optimization problem (3.1), under the condition that $\boldsymbol{\beta}^*$ is a feasible solution, the optimality of $\widehat{\boldsymbol{\beta}}^{(l)}$ yields

$$\|\boldsymbol{\beta}_S^*\|_1 = \|\boldsymbol{\beta}^*\|_1 \geq \|\widehat{\boldsymbol{\beta}}^{(l)}\|_1 = \|\widehat{\boldsymbol{\beta}}_S^{(l)}\|_1 + \|\widehat{\boldsymbol{\beta}}_{S^c}^{(l)}\|_1 = \|\widehat{\boldsymbol{\beta}}_S^{(l)}\|_1 + \|(\widehat{\boldsymbol{\beta}}^{(l)} - \boldsymbol{\beta}^*)_{S^c}\|_1, \quad \text{(D.1)}$$

where the last equality follows from the fact that $(\boldsymbol{\beta}^*)_{S^c} = \mathbf{0}$. (D.1) immediately leads to

$$\|\boldsymbol{\beta}_S^*\|_1 - \|\widehat{\boldsymbol{\beta}}_S^{(l)}\|_1 \geq \|(\widehat{\boldsymbol{\beta}}^{(l)} - \boldsymbol{\beta}^*)_{S^c}\|_1.$$

Moreover, by triangle inequality, we have

$$\|(\widehat{\boldsymbol{\beta}}^{(l)} - \boldsymbol{\beta}^*)_S\|_1 \geq \|\boldsymbol{\beta}_S^*\|_1 - \|\widehat{\boldsymbol{\beta}}_S^{(l)}\|_1.$$

Combining the above two inequalities, we can obtain

$$\|(\widehat{\boldsymbol{\beta}}^{(l)} - \boldsymbol{\beta}^*)_S\|_1 \geq \|(\widehat{\boldsymbol{\beta}}^{(l)} - \boldsymbol{\beta}^*)_{S^c}\|_1.$$

This completes the proof. $\square$

## D.2   Proof of Lemma C.2

*Proof.* In the optimization problem (3.3), under the condition that $\boldsymbol{\theta}_j^*$ is a feasible solution, the optimality of $\widehat{\boldsymbol{\theta}}_j^{(l)}$ yields

$$\|(\boldsymbol{\theta}_j^*)_{S_{\theta_j}}\|_1 = \|\boldsymbol{\theta}_j^*\|_1 \geq \|\widehat{\boldsymbol{\theta}}_j^{(l)}\|_1 = \|(\widehat{\boldsymbol{\theta}}_j^{(l)})_{S_{\theta_j}}\|_1 + \|(\widehat{\boldsymbol{\theta}}_j^{(l)})_{S_{\theta_j}^c}\|_1 = \|(\widehat{\boldsymbol{\theta}}_j^{(l)})_{S_{\theta_j}}\|_1 + \|(\widehat{\boldsymbol{\theta}}_j^{(l)} - \boldsymbol{\theta}_j^*)_{S_{\theta_j}^c}\|_1, \quad \text{(D.2)}$$

where the last equality follows from the fact that $(\boldsymbol{\theta}_j^*)_{S_{\theta_j}^c} = \mathbf{0}$. (D.2) immediately leads to

$$\|(\boldsymbol{\theta}_j^*)_{S_{\theta_j}}\|_1 - \|(\widehat{\boldsymbol{\theta}}_j^{(l)})_{S_{\theta_j}}\|_1 \geq \|(\widehat{\boldsymbol{\theta}}_j^{(l)} - \boldsymbol{\theta}_j^*)_{S_{\theta_j}^c}\|_1. \quad \text{(D.3)}$$

Moreover, by triangle inequality, we have

$$\|(\widehat{\boldsymbol{\theta}}_j^{(l)} - \boldsymbol{\theta}_j^*)_{S_{\theta_j}}\|_1 \geq \|(\boldsymbol{\theta}_j^*)_{S_{\theta_j}}\|_1 - \|(\widehat{\boldsymbol{\theta}}_j^{(l)})_{S_{\theta_j}}\|_1. \quad \text{(D.4)}$$

Combining (D.3) and (D.4), we can obtain

$$\|(\widehat{\boldsymbol{\theta}}_j^{(l)} - \boldsymbol{\theta}_j^*)_{S_{\theta_j}}\|_1 \geq \|(\widehat{\boldsymbol{\theta}}_j^{(l)} - \boldsymbol{\theta}_j^*)_{S_{\theta_j}^c}\|_1.$$

This completes the proof. $\square$

# E   Auxiliary Definitions, Lemmas and Theorems

We define sub-Exponential random variables and its corresponding $\psi_1$ norm as follows.

**Definition E.1** (Definition 5.13 in Vershynin (2010)). *A random variable $X$ is called sub-Exponential if there exists a constant $K > 0$ such that for all $p \geq 1$ the following inequality holds:*

$$(\mathbb{E}(|X|^p))^{1/p} \leq Kp. \tag{E.1}$$

*The $\psi_1$ norm of $X$, denoted as $\|X\|_{\psi_1}$, is the smallest $K$ that makes (E.1) holds. In other words,*

$$\|X\|_{\psi_1} = \sup_{p \geq 1} p^{-1} (\mathbb{E}(|X|^p))^{1/p}.$$

Similarly, sub-Gaussian random variables and the corresponding $\psi_2$ norm are defined as follows:

**Definition E.2** (Definition 5.7 in Vershynin (2010)). *A random variable $X$ is called sub-Gaussian if there exists a constant $K > 0$ such that for all $p \geq 1$ the following inequality holds:*

$$(\mathbb{E}(|X|^p))^{1/p} \leq K\sqrt{p}. \tag{E.2}$$

*The $\psi_2$ norm of $X$, denoted as $\|X\|_{\psi_2}$, is the smallest $K$ that makes (E.2) holds. In other words,*

$$\|X\|_{\psi_2} = \sup_{p \geq 1} p^{-1/2} (\mathbb{E}(|X|^p))^{1/p}.$$

We can generalize the concept of sub-Gaussian random variable to sub-Gaussian random vector.

**Definition E.3** (Definition 5.22 in Vershynin (2010)). *A random vector $\boldsymbol{X} \in \mathbb{R}^d$ is sub-Gaussian if for any vector $\mathbf{u} \in \mathbb{R}^d$ the inner product $\langle \boldsymbol{X}, \mathbf{u} \rangle$ is a sub-Gaussian random variable. And the corresponding $\psi_2$ norm of $\boldsymbol{X}$ is defined as*

$$\|\boldsymbol{X}\|_{\psi_2} = \sup_{\|\mathbf{u}\|_2 = 1} \|\langle \boldsymbol{X}, \mathbf{u} \rangle\|_{\psi_2}.$$

It is obvious that for any sub-Gaussian random vector $\boldsymbol{X} \in \mathbb{R}^d$, $\|\boldsymbol{X}\|_{\psi_2} \geq \max_{j=1}^d \|X_j\|_{\psi_2}$.

It is proved in Vershynin (2010) that a centered Gaussian random variable $X$ with variance $\sigma^2$ is also a sub-Gaussian random variable with $\|X\|_{\psi_2} \leq C\sigma$ where $C$ is an absolute constant. Therefore, we can easily show that a centered Gaussian random vector $\boldsymbol{X}$ with covariance matrix $\boldsymbol{\Sigma}$ is also a sub-Gaussian random vector with $\|\boldsymbol{X}\|_{\psi_2} \leq C\lambda_{\max}(\boldsymbol{\Sigma})$.

The following theorem is the Hoeffding type inequality for sub-Gaussian random variables, which characterizes the tail bound for the weighted sum of independent sub-Gaussian random variables.

**Theorem E.4** (Proposition 5.10 in Vershynin (2010)). *Let $X_1, X_2, \ldots, X_n$ be independent centered sub-Gaussian random variables, and let $K = \max_i \|X_i\|_{\psi_2}$. Then for every $\mathbf{a} = (a_1, a_2, \ldots, a_n) \in \mathbb{R}^n$ and for every $t > 0$, we have*

$$\mathbb{P}\bigg(\bigg|\sum_{i=1}^n a_i X_i\bigg| > t\bigg) \leq \exp\bigg(-\frac{Ct^2}{K^2 \|\mathbf{a}\|_2^2}\bigg),$$

*where $C > 0$ is an absolute constant.*

Similarly, the following theorem is the Bernstein type inequality for sub-Exponential random variables, which provides the tail bound on the weighed sum of independent sub-Exponential random variables.

**Theorem E.5** (Proposition 5.16 in Vershynin (2010)). *Let $X_1, X_2, \ldots, X_n$ be independent centered sub-Exponential random variables, and let $K = \max_i \|X_i\|_{\psi_1}$. Then for every $\mathbf{a} = (a_1, a_2, \ldots, a_n) \in \mathbb{R}^n$ and for every $t > 0$, we have*

$$\mathbb{P}\bigg(\bigg|\sum_{i=1}^n a_i X_i\bigg| > t\bigg) \leq 2\exp\bigg(-C\min\bigg\{\frac{t^2}{K^2\|\mathbf{a}\|_2^2}, \frac{t}{K\|\mathbf{a}\|_\infty}\bigg\}\bigg),$$

*where $C > 0$ is an absolute constant.*

**Lemma E.6.** *For $X_1$ and $X_2$ being two sub-Gaussian random variables, $X_1 X_2$ is a sub-Exponential random variable with*

$$\|X_1 X_2\|_{\psi_1} \leq C \max\{\|X_1\|_{\psi_2}^2, \|X_2\|_{\psi_2}^2\},$$

*where $C > 0$ is an absolute constant.*

Lemma E.6 reveals that the product of two sub-Gaussian random variables is a sub-Exponential random variable and gives an upper bound on its $\psi_1$ norm.

**Theorem E.7** (Corollary 1 in Raskutti et al. (2010)). *For any design matrix $\mathbf{X} \in \mathbb{R}^{n \times d}$ with i.i.d. rows following Gaussian distribution $N(0, \boldsymbol{\Sigma})$, if $\boldsymbol{\Sigma}$ satisfies the RE condition with parameter $(s, \alpha, \gamma)$, and the sample size $n$ satisfies*

$$n > \frac{C''\rho^2(\boldsymbol{\Sigma})(1+\alpha)^2}{\gamma^2} s \log d,$$

*then the sample covariance matrix $\widehat{\boldsymbol{\Sigma}} = \mathbf{X}^\top \mathbf{X}/n$ satisfies the restricted eigenvalue condition with parameter $(s, \alpha, \gamma/8)$ with probability at least $1 - C'\exp(-Cn)$, where $C$, $C'$ and $C''$ are absolute constants and $\rho^2(\boldsymbol{\Sigma}) = \max_{1 \leq j \leq d} \Sigma_{jj}$.*